\documentclass[letterpaper]{article} 
\usepackage{aaai25}  
\usepackage{times}  
\usepackage{helvet}  
\usepackage{courier}  
\usepackage[hyphens]{url}  
\usepackage{graphicx} 
\usepackage{natbib}  
\usepackage{caption} 
\usepackage{algorithm}

\usepackage{textcomp}
\usepackage{url}
\usepackage{verbatim}
\usepackage{graphicx}
\usepackage{tikz}
\usepackage{comment}
\usepackage{color}

\usepackage{amsmath}
\usepackage{amssymb}
\usepackage{booktabs}
\usepackage{mathtools}
\usepackage{multirow, array,makecell,xspace,bm}
\usepackage{pifont}
\usepackage{amsfonts}
\usepackage{arydshln}
\usepackage{dsfont}

\usepackage{bibentry}
\usepackage{algpseudocode}
\usepackage{stackengine}

\usepackage{scalerel}
\usepackage{esdiff}

\newcommand{\cmark}{\ding{51}}%
\usepackage{newfloat}
\usepackage{listings}
\DeclareCaptionStyle{ruled}{labelfont=normalfont,labelsep=colon,strut=off} 
\DeclareMathOperator*{\argmax}{arg\,max}

\usepackage{newfloat}
\usepackage{listings}
\DeclareCaptionStyle{ruled}{labelfont=normalfont,labelsep=colon,strut=off} 
\floatstyle{ruled}
\newfloat{listing}{tb}{lst}{}
\floatname{listing}{Listing}
\title{Multispectral Pedestrian Detection with Sparsely Annotated Label}

\author{Chan Lee\thanks{Equal contribution}, Seungho Shin$\footnotemark[1]$, 
Gyeong-Moon Park\thanks{Corresponding author}, Jung Uk Kim$\footnotemark[2]$
}
\affiliations {
    Kyung Hee University, Yong-in, South Korea\\
    \{cksdlakstp12, ssh9918, gmpark, ju.kim\}@khu.ac.kr
}

\usepackage{bibentry}

\begin{document}

\maketitle

\begin{abstract}
    Although existing Sparsely Annotated Object Detection (SAOD) approaches have made progress in handling sparsely annotated environments in multispectral domain, where only some pedestrians are annotated, they still have the following limitations: (\textit{i}) they lack considerations for improving the quality of pseudo-labels for missing annotations, and (\textit{ii}) they rely on fixed ground truth annotations, which leads to learning only a limited range of pedestrian visual appearances in the multispectral domain. To address these issues, we propose a novel framework called Sparsely Annotated Multispectral Pedestrian Detection (SAMPD). For limitation (\textit{i}), we introduce Multispectral Pedestrian-aware Adaptive Weight (MPAW) and Positive Pseudo-label Enhancement (PPE) module. Utilizing multispectral knowledge, these modules ensure the generation of high-quality pseudo-labels and enable effective learning by increasing weights for high-quality pseudo-labels based on modality characteristics. To address limitation (\textit{ii}), we propose an Adaptive Pedestrian Retrieval Augmentation (APRA) module, which adaptively incorporates pedestrian patches from ground-truth and dynamically integrates high-quality pseudo-labels with the ground-truth, facilitating a more diverse learning pool of pedestrians. Extensive experimental results demonstrate that our SAMPD significantly enhances performance in sparsely annotated environments within the multispectral domain. The code is available at \url{https://github.com/VisualAIKHU/SAMPD}.
\end{abstract}

\section{Introduction}

Recently, multispectral pedestrian detection has gained attention in computer vision \cite{chen2022multimodal,chen2023attentive, kim2024causal}. Unlike single-modal detectors that use only one modality (\textit{e.g.,} visible or thermal), multispectral pedestrian detection combines visible and thermal images \cite{kim2021robust, kim2023similarity}. Visible images capture texture and color, while thermal images provide heat signatures. Combining the two modalities enhances detection robustness across various conditions, including low-light environments \cite{jia2021llvip, 9706418} and adverse weather conditions \cite{hwang2015multispectral}.

\begin{figure}[t]
    \begin{minipage}[b]{0.999\linewidth}
	\centering
        \centerline{\includegraphics[width=\linewidth]{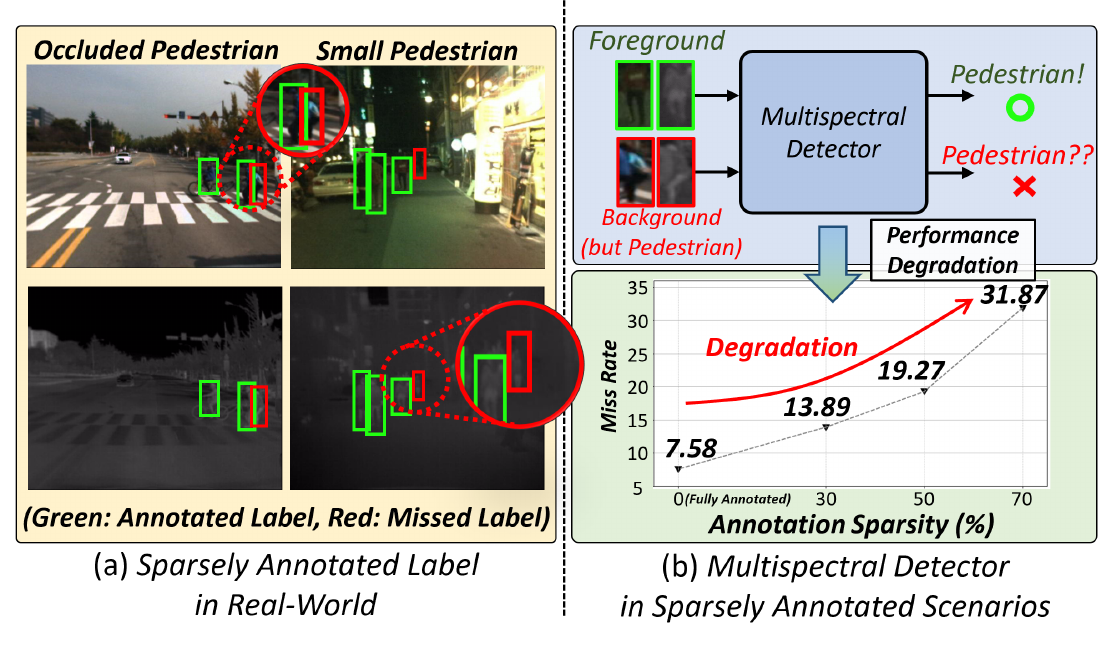}}
        \end{minipage}
	\caption {(a) Examples of sparsely annotated labels in multispectral domain. (b) In sparse annotation scenarios, multispectral pedestrian detectors struggle to learn pedestrians.}
    \label{fig:1}
    \vspace{-0.35cm}
\end{figure}

However, multispectral pedestrian detection faces challenges due to frequently occurring sparse annotations, often caused by human errors in annotating small or occluded pedestrians, even when they are present \cite{li2019illumination}. In sparsely annotated environments, a real pedestrian might be annotated as a pedestrian in some instances but not in others (Figure \ref{fig:1}(a)). This inconsistency causes the network to struggle with effectively learning knowledge from both modalities, resulting in a significant decline in pedestrian detection performance (Figure \ref{fig:1}(b)). For this reason, while some existing multispectral datasets \cite{hwang2015multispectral, jia2021llvip} have reached near-perfect annotation at high labor and time costs, they may still have gaps, making robust research in sparsely annotated environments essential.

To address the above-mentioned issue, Sparsely Annotated Object Detection (SAOD) task has been introduced \cite{niitani2019sampling,zhang2020solving,wang2021comining,wang2023calibratedteachersparselyannotated,suri2023sparsedet}. The task aims to generate pseudo-labels to address missing annotations. Current methods include selecting high-confidence boxes from model predictions \cite{niitani2019sampling, wang2023calibratedteachersparselyannotated}, adjusting loss functions to incorporate both original and augmented images \cite{wang2021comining}, and applying self-supervised loss to avoid negative gradient propagation \cite{suri2023sparsedet}. However, these methods often assume pseudo-labels are reliable, even when they are inaccurate or do not fully capture pedestrian appearance. Furthermore, reliance on a fixed set of ground-truth annotations makes it challenging to incorporate valuable missing annotations, limiting the ability to learn about diverse pedestrian visual appearances.

In this paper, we present a novel method, Sparsely Annotated Multispectral Pedestrian Detection (SAMPD), to tackle the challenges of sparse annotation in the multispectral domain. Our approach considers two main challenges: (\textit{i}) how to effectively learn multispectral pedestrian information from pseudo-labels and enhance their quality, and (\textit{ii}) how to integrate the identified missing annotations during training to enable more comprehensive learning.

To address the challenge (\textit{i}), we first adopt a teacher-student structure, as used in existing SAOD works. However, we newly introduce two modules: Multispectral Pedestrian-aware Adaptive Weight (MPAW) and Positive Pseudo-label Enhancement (PPE). The MPAW module aims to help the student model learn multispectral modalities by assigning higher weights to high-quality pseudo-labels based on modality characteristics (single and multispectral). The PPE module aligns feature representations to make high-quality pseudo-labels more similar to each other, while distancing them from low-quality pseudo-labels. We also consider the uncertain pseudo-labels to prevent them from misguiding the model. This approach allows the student model to learn from more reliable pseudo-labels, thereby improving robustness in environments with sparse annotations.

For the challenge (\textit{ii}), we propose an Adaptive Pedestrian Retrieval Augmentation (APRA) module. In sparsely annotated environments, finding missing annotations by leveraging diverse visual representations of pedestrians is crucial. The APRA module adaptively attaches ground-truth pedestrian patches that best match the input image based on lighting conditions. Additionally, high-quality pseudo-labels are dynamically integrated with the ground-truth. This dynamic augmentation approach enriches the visual representation of pedestrians, allowing the teacher model to generate more reliable pseudo-labels and enhancing the robustness of the student model, which is our final detector during inference.

As a result, SAMPD effectively addresses sparsely annotated settings by generating more reliable pseudo-labels. It also improves performance in fully annotated scenarios, as real-world images may still have missing annotations.

The major contributions can be summarized as follows: 
\begin{itemize}
\item We propose the Multispectral Pedestrian-aware Adaptive Weight (MPAW) module to adjust the learning of each modality for improved pseudo-label learning.

\item We introduce the Positive Pseudo-label Enhancement (PPE) module to help our SAMPD generate higher-quality pseudo-labels. 

\item We develop the Adaptive Pedestrian Retrieval Augmentation (APRA) module to enrich pedestrian knowledge by adaptively augmenting images and integrating high-quality pseudo-labels into ground-truth labels.

\end{itemize}

\section{Related Work}
\subsection{Multispectral Pedestrian Detection}

Multispectral pedestrian detection has gained attention for its impressive performance \cite{park2022robust, 10222560, e25071022, zhu2023dpacfuse}. AR-CNN tackles modality alignment issues \cite{zhang2019weakly}, while IATDNN+IASS introduces illumination-aware mechanisms for robustness \cite{guan2019fusion}. MBNet and AANet address modal discrepancy \cite{zhou2020improving, chen2023attentive}, and MLPD proposes multi-label and non-paired augmentation \cite{kim2021mlpd}. ProbEn adopts late-fusion with probability ensembling \cite{chen2022multimodal}, and DCMNet uses local and non-local aggregation for contextual information \cite{xie2022learning}. Liu et al. \cite{liu2024region} improves performance by utilizing illumination and temperature information. Beyond Fusion \cite{xie2024beyond} introduces a hallucination branch to map thermal to visible domains.

While these studies have shown promising detection performances, they assume perfect bounding box annotations \cite{kim2021uncertainty, kim2021robust, kim2023similarity}, which may not be realistic in real-world scenarios. Challenges such as differences between visible and thermal images, small or occluded pedestrians, and human error can lead to imperfect (\textit{i.e.,} sparse) annotations. Existing methods struggle in such scenarios. To address this, we introduce Sparse Annotation Multispectral Pedestrian Detection (SAMPD), which fully leverages multispectral knowledge to effectively generate and enhance pseudo-labels even with sparse annotations.

\subsection{Sparsely Annotated Object Detection}
When pedestrian annotations are incomplete, it can lead to inconsistencies where pedestrians are misclassified as either foreground or background. To tackle this issue, Sparsely Annotated Object Detection (SAOD) has introduced. Pseudo label methods \cite{niitani2019sampling} establish logical connections between object co-occurrence and pseudo-label application. BRL \cite{zhang2020solving} introduces automatic adjustments for areas prone to mislabeling. Co-mining \cite{wang2021comining} jointly trains models using predictions from both original and augmented images. SparseDet \cite{suri2023sparsedet} introduces a self-supervised loss to prevent negative gradient propagation. Calibrated Teacher \cite{wang2023calibratedteachersparselyannotated} validates pseudo-label quality using a calibrator that distinguishes positive from negative labels. However, existing methods rely on the pseudo-labels without considering their quality. In contrast, our SAMPD enhances pseudo-label quality by integrating multispectral knowledge and incorporating filtered high-quality pseudo-labels through the Adaptive Pedestrian Retrieval Augmentation (APRA) module.

\begin{figure*}[t] 
  \centering
  \includegraphics[width=0.95\linewidth]{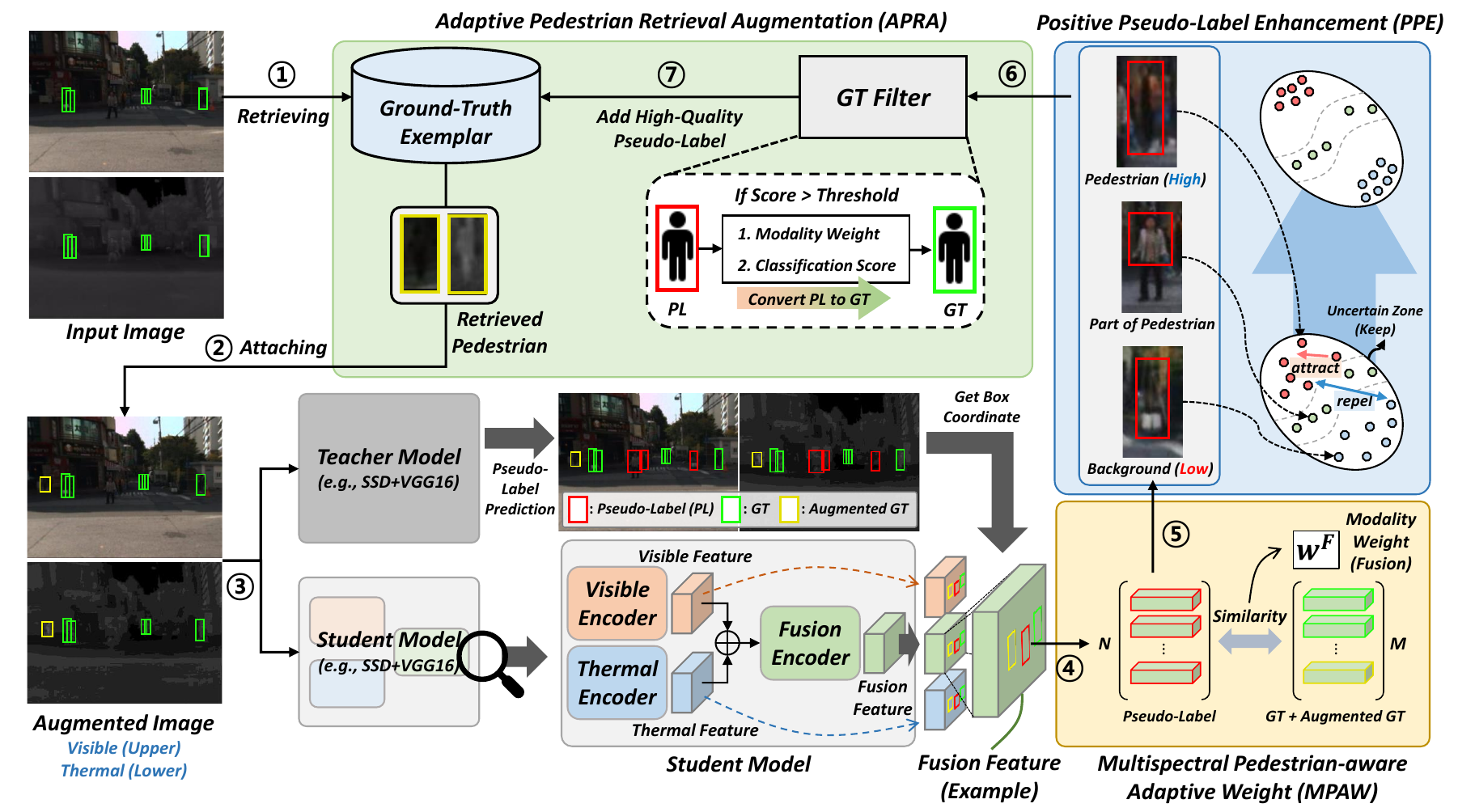}
  \caption{Overall architecture of our SAMPD. Given the multispectral input image, the APRA module generates an augmented image, which is then used as the input for the training phase. $\boldsymbol{\oplus}$ denotes the concatenation. During inference phase, only student model is used for detection, and only the original input image (without augmentation) is utilized.}
  \label{fig:2}
  \vspace{-0.35cm}
\end{figure*}

\subsection{Retrieval-Augmented Generation}
Retrieval-Augmented Generation (RAG) has proven to enhance the accuracy and reliability of generative AI, addressing the challenge of limited access to the latest information after training phase. Initially developed for natural language processing (NLP) \cite{lewis2021retrievalaugmentedgenerationknowledgeintensivenlp}, RAG has proven effective in various applications, including image captioning \cite{ramos2023retrievalaugmentedimagecaptioning, sarto2022retrievalaugmentedtransformerimagecaptioning}, multimodal learning. Traditionally, it compensates for the lack of information in high-accuracy modalities by storing features of that modality in memory. We propose a novel RAG method, \textit{i.e.,} APRA module, for effective multispectral pedestrian detection. By leveraging the unique characteristics of multispectral data, our APRA module enables the learning of diverse pedestrian information. It helps create a teacher model that generates more reliable pseudo-labels, improving the overall accuracy of pedestrian detection.

\section{Proposed Method}


Figure \ref{fig:2} shows the overall architecture of our SAMPD in the training phase. The APRA module retrieves $m$ pedestrian patches to create augmented image pair. The teacher and student model receive an augmented image pair and encode through each modality backbone. Then, the teacher model generates pseudo-labels and passes them to the student model. The student model employs the MPAW module that assigning higher weights to high-quality pseudo-labels. The PPE module guides the pseudo-labels in the feature space to distinguish the high and low quality pseudo-labels. Finally, the APRA enriches pedestrian knowledge by adaptively augmenting images and integrating high-quality pseudo-labels into the ground-truth annotation. 

\subsection{Multispectral Pedestrian-aware Adaptive Weight}

As shown in Figure \ref{fig:2}, we adopt a teacher-student structure commonly used in SAOD methods. The student model, our final detector, learns from pseudo-labels generated by the teacher model. The multispectral data comprises visible ($V$), thermal ($T$), and fusion ($F$) modalities. To generate effective pseudo-labels in sparsely annotated scenarios, both the teacher and student models are constructed with 3-way encoding paths, fully exploiting single-modal ($V$ and $T$) and multispectral ($F$) knowledge during training.

However, some pseudo-labels from the teacher model may be incorrect, capturing only parts of pedestrians or including background elements. These errors can negatively impact the performance of the student model and cause confusion during training. To address this, we propose a multispectral pedestrian-aware adaptive weight (MPAW) module to increase the influence of reliable pseudo-labels and reduce the impact of unreliable ones for each modality. This approach emphasizes high-quality pseudo-labels and reliable modalities. For each modality $k=\{V,T,F\}$, we extract $N$ feature maps from the pseudo-label boxes of the student model $\textbf{f}_{PL}^{k(s)}=\{f_{PL_i}^{k(s)}\}_{i=1}^{N}$ and $M$ feature maps from the ground-truth labels $\textbf{f}_{GT}^{k(s)}=\{f_{GT_i}^{k(s)}\}_{i=1}^{M}$. We then apply global average pooling (GAP) to obtain latent vectors $\textbf{l}_{PL}^{k(s)}=\{l_{PL_i}^{k(s)}\}_{i=1}^{N}$ and $\textbf{l}_{GT}^{k(s)}=\{l_{GT_i}^{k(s)}\}_{i=1}^{M}$. We compute the modality weight for $k$ modality, $w^k$, as defined by:
\begin{gather}
    w^k = \frac{1}{N}{\sum_{i=1}^{N} \max_{j\in\{1,...,M\}}d(l_{PL_i}^{k(s)}, l_{GT_j}^{k(s)}}), \\
    d(\alpha,\beta)=\frac{\alpha \cdot \beta} {{||\alpha||}\,{||\beta||}},
\end{gather}
where $N$ and $M$ denote the number of pseudo-labels and ground-truth labels, respectively, and $d(\cdot,\cdot)$ is the cosine similarity. We compute the maximum cosine similarity between $l_{PL}^{k(s)}$ and the closest vector in $l_{GT}^{k(s)}$. A high $w^k$ indicates that the pseudo-labels for modality $k$ closely match the ground-truth labels, reflecting reliable quality, while a low $w^k$ signifies less reliable pseudo-labels.

Using $w^k$, the detection losses $\mathcal{L}_{det}^{sum}$ from the 3-way encoding paths are computed as follows:
\begin{equation}
        \mathcal{L}_{det}^{sum} = w^{V}\mathcal{L}_{det}^{V(s)} + w^{T}\mathcal{L}_{det}^{T(s)} + w^{F}\mathcal{L}_{det}^{F(s)},
    \label{eq3}
\end{equation}
where $\mathcal{L}_{det}^{k(s)}$ represents the detection loss \cite{kim2021mlpd,kim2021robust} of the student model, including the classification loss $\mathcal{L}_{cls}^{k(s)}$ and the localization loss $\mathcal{L}_{loc}^{k(s)}$ for $k$ modality. By applying Eq. (\ref{eq3}), the teacher model focuses on reliable pseudo-labels and modalities, enabling more stable learning in sparsely annotated scenarios.

\subsection{Positive Pseudo-label Enhancement}
Through the MPAW module, the student model can learn from the pseudo-labels. In addition, we introduce a Positive Pseudo-label Enhancement (PPE) module to enable the teacher model to generate higher-quality pseudo-labels. With $\textbf{l}_{PL}^{k(s)}$ and $\textbf{l}_{GT}^{k(s)}$ from student model of $k$ modality ($k={V,T,F}$), we calculate the cosine similarity $d(\cdot,\cdot)$ to construct the \textit{positive} (foreground) and \textit{negative} (background) pseudo-labels, which can be represented as:
\begin{align}
    a=\argmax_{j\in\{1,...,M\}}d(l_{PL_i}^{k(s)}, l_{GT_j}^{k(s)}).
    \label{eq:4}
\end{align}
We classify the $i$-th pseudo-label as \textit{positive} pseudo-label, if the similarity between the $i$-th pseudo-label and the most similar ground-truth label exceeds the $\tau_1$. Conversely, if the similarity is below $\tau_2$, $i$-th pseudo-label is classified as  \textit{negative} pseudo-label. Also, the pseudo-labels with similarity scores between $\tau_1$ and $\tau_2$ fall into an \textit{uncertain} pseudo-label, acting as a buffer to avoid bias towards either \textit{positive} or \textit{negative}, thereby improving the reliability of the learning process. We set $\tau_1=0.9$ and $\tau_2=0.7$, respectively.

After the distinction, we devise a positive pseudo-label guiding (PG) loss for the $k$ modality $\mathcal{L}_{PG}^k$, which can be represented as:
\begin{gather}
    p_{PL}(p_i) = \sum_{j=1}^{N_p}\exp{(d(l_{PL_{p_i}}^{k(s)}, l_{PL_{p_j}}^{k(s)})/\tau)},\\
    n_{PL}(p_i) = \sum_{j=1}^{N_n}\exp{(d(l_{PL_{p_i}}^{k(s)}, l_{PL_{n_j}}^{k(s)})/\tau)},\\
    \mathcal{L}_{PG}^k = -\frac{1}{N_p}\sum_{i=1}^{N_p}\log\frac{p_{PL}(p_i)}{p_{PL}(p_i)+n_{PL}(p_i)},
    \label{eq6}
\end{gather}
where $p$ and $n$ indicates \textit{positive} and \textit{negative} pseudo-labels, $N_p$ and $N_n$ is the number of \textit{positive} and \textit{negative} pseudo-labels, and $\tau$ is the temperature parameter. 

The purpose of the PG loss is to use the differentiated \textit{positive} and \textit{negative} labels from the student model to guide the feature representations of \textit{positive} pseudo-labels, promoting those similar to the ground-truth labels provided by the teacher model to move closer together. This process helps train the teacher model to generate higher-quality pseudo-labels. Since the PG loss is applied across visible, thermal, and fusion modalities, it leverages diverse multimodal knowledge, enhancing learning stability.

\begin{figure}[t] 
  \centering
  \includegraphics[width=\linewidth]{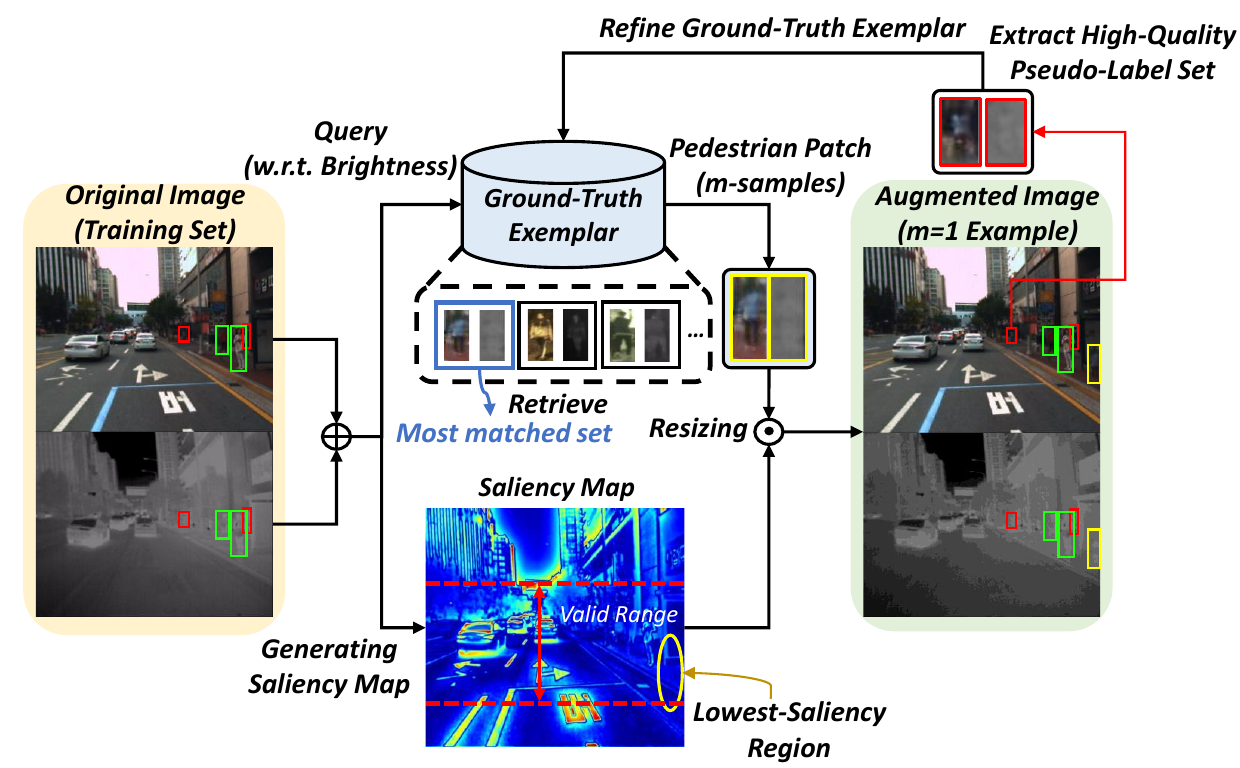}
      \caption{Given an original image, our adaptive pedestrian retrieval augmentation (APRA) module finds $m$ pedestrian patches from a ground-truth exemplar with similar brightness to the original image ($m=1$ example). The APRA module uses the saliency map of the original image to locate the region with the lowest saliency value and attaches the resized $m$ pedestrian patches, according to the ground-truth bounding-box size, to this region. In addition, if the pseudo-labels are considered reliable, the corresponding pedestrian patch is included in the ground-truth exemplar and used as ground-truth in the subsequent training process. $\oplus$ denotes the concatenation, while $\odot$ indicates the attaching process.}
  \label{fig:3}
\end{figure}
    
\begin{table*}[t]
    \renewcommand{\tabcolsep}{3.5mm}
    \centering
    \resizebox{0.94\linewidth}{!}{
    \begin{tabular}{c c c c c c c c c c c c c}
        \Xhline{3\arrayrulewidth}
    \rule{0pt}{10pt}\multirow{2}{*}{\bf Methods} & \multicolumn{3}{c}{\bf 30\%} & \multicolumn{3}{c}{\bf 50\%} & \multicolumn{3}{c}{\bf 70\%}\\ \cmidrule(lr){2-4} \cmidrule(lr){5-7} \cmidrule(lr){8-10}
        &\bf  All         &\bf  Day         & \bf Night       & \bf All         & \bf Day         & \bf Night      & \bf All         & \bf Day         & \bf Night\\ \hline
        \rule{0pt}{9.0pt}
        Supervised  & 13.89           & 15.50           & 10.88           & 19.27           & 21.16           & 15.66           & 31.87          &  33.73          & 27.81  \\ \cdashline{1-13}             
        \rule{0pt}{9.0pt}
        Pseudo label (CVPR'19)                 & 11.95           & 14.09           & 8.14           & 18.09           & 21.00           & 13.33           & 29.43          &  31.24          & 26.00 \\
        BRL (ICASSP'20)      & 11.90           & 13.61           & 8.69           & 17.93           & 19.62           & 14.42           & 28.71          &  32.10          & 21.48       \\
        Co-mining (AAAI'21)        & 11.66           & 12.68           & 9.80           & 17.90           & 19.43           & 14.91           & 28.80          &  28.48          & 29.41        \\
        SparseDet (ICCV'23)                & 10.92           & 11.99           & 8.78           & 18.07           & \underline{19.30}           & 15.40          & 28.20          &  28.80          & 26.54    \\
        Calibrated Teacher (AAAI'23)                & \underline{10.47}           & \underline{11.81}           & \underline{7.82}           & \underline{17.67}           & 19.77           & \underline{13.19}          & \underline{25.48}          &  \underline{28.43}          & \underline{19.28}    \\ \cdashline{1-13}
        \rule{0pt}{9.0pt} \bf SAMPD (Ours) & \bf 8.56 & \bf 10.55 & \bf 5.62           & \bf 15.27           & \bf 17.28           & \bf 11.15           & \bf 23.52          &  \bf 26.15          & \bf 17.87 
        \\\Xhline{3\arrayrulewidth}
    \end{tabular}
}
\caption{Sparsely annotated detection results (MR) on the KAIST dataset by varying removal percentages (30\%, 50\%, and 70\%) for sparsely annotated scenarios. We compared our method with state-of-the-art SAOD methods that address sparsely annotated scenario. \textbf{Bold}/\underline{underlined} fonts indicate the best/second-best results.}
\vspace{-0.34cm}
\label{table:KAIST}
\end{table*}

\subsection{Adaptive Pedestrian Retrieval Augmentation}
While MPAW and PPE modules effectively utilize pseudo-labels, there remains a challenge due to the limited visual diversity in ground-truth annotations within sparsely annotated environments. To address this, we propose an Adaptive Pedestrian Retrieval Augmentation (APRA) module that adaptively integrates additional pedestrian patches into the original image to enhance the ground-truth information. \\

\noindent\textbf{Augmenting Images with Pedestrian Patches.} Figure \ref{fig:3} shows the operation of the APRA module. A ground-truth exemplar retrieves relevant pedestrian patches by selecting $m$ patches with the smallest brightness difference from the original image. These patches are resized to the average annotation size for each image before being integrated into the image. If no annotations are present, the patches are resized to the average size of all annotations before being attached.

To ensure effective integration of these patches, we carefully select regions within the image, avoiding areas with a low likelihood of containing pedestrians, such as the sky or directly in front of vehicles. We define valid pedestrian locations using the mean, variance, and standard deviation of the \textit{y}-coordinates from existing annotations within a 90\% confidence interval. A saliency map identifies regions with a low probability of containing pedestrians, targeting these areas for patch integration. By focusing on low-saliency regions, the APRA module enhances the ground-truth representation, leading to improved performance.\\

\noindent\textbf{Dynamic Ground-Truth Refinement.} Additionally, we refine ground-truth annotations by dynamically converting pseudo-labels into ground-truth labels. This process uses $w^k$ ($k=V,T,F$) from Eq. (1) and the classification score of the pseudo-label. If both values exceed the threshold $\tau_1$ (indicating high-quality pseudo-labels in the PPE module), the pseudo-label is converted into a ground-truth label. These converted labels are then integrated into the ground-truth annotations, ensuring consistent recognition of reliable pseudo-labels and improving model stability. Through the APRA module, our framework achieves more reliable annotations, reducing gaps in ground-truth information within sparsely annotated environments.

\subsection{Total Loss}
The total loss function of our SAMPD is represented as:
\begin{equation}
    \mathcal{L}_{\textit{Total}} = \lambda_1\mathcal{L}_{det}^{sum} + \lambda_2(\mathcal{L}_{PG}^V + \mathcal{L}_{PG}^T + \mathcal{L}_{PG}^F),
    \label{eq7}
\end{equation}

\noindent where $\mathcal{L}_{det}^{sum}$ denotes the detection loss using. $\lambda_1$ and $\lambda_2$ denote the balancing hyper-parameters. Using the $\mathcal{L}_{\textit{Total}}$, our SAMPD shows the robust detection performance even in the sparsely annotation scenarios.

\section{Experiments}

\subsection{Dataset and Evaluation Metric}

\noindent\textbf{KAIST Dataset.} The KAIST dataset \cite{hwang2015multispectral} comprises 95,328 pairs of visible and thermal images, enriched with 103,128 bounding box annotations to identify pedestrians. We use a test set of 2,252 images to evaluate performance. Following \cite{suri2023sparsedet}, we simulated a sparsely annotated scenario by increasing the probability of removing bounding-box annotations with smaller widths from the training set. More details are in the supplementary document. Finally, we removed 30\%, 50\%, and 70\% of the bounding-box annotations among the total annotations, consistent with the ratios described in \cite{niitani2019sampling}. \\

\noindent\textbf{LLVIP Dataset.} The LLVIP dataset \cite{jia2021llvip} contains visible-thermal paired dataset for low-light vision. It consists 15,488 visible-thermal image pairs. Following the same protocol as the KAIST dataset, we simulated a sparsely annotated scenario for the LLVIP dataset by removing 30\%, 50\%, and 70\% of the bounding-box annotations from the total annotations. \\

\begin{table}[t!]
    \renewcommand{\tabcolsep}{1.2mm}
	\centering
        \resizebox{1.0\linewidth}{!}
		{
	\begin{tabular}{c ccc ccc}
            \Xhline{3\arrayrulewidth}
            \rule{0pt}{10pt} \multirow{2}{*}{\bf Methods} & \multicolumn{3}{c}{\bf MR} & \multicolumn{3}{c}{\bf AP$_{50}$} \\ 
            \cmidrule(lr){2-4} \cmidrule(lr){5-7}
            &\bf  30\%         &\bf  50\%         & \bf 70\%       &\bf  30\%         &\bf  50\%         & \bf 70\%
            
            \\ \hline
            \rule{0pt}{10.0pt}
            Supervised & 11.87 & 14.25 & 17.77 & 92.31 & 89.78 & 87.11 \\ 
            \cdashline{1-7}
            \rule{0pt}{10.2pt} 
            Pseudo label (CVPR'19) & 10.91 & 13.21 & 17.45 & 93.85 & 92.41 & 89.49 \\ 
            BRL (ICASSP'20) & 10.93 & 12.74 & 17.24 & 94.46 & 93.37 & 90.89 \\
            Co-mining (AAAI'21) & 10.88 & 12.89 & 16.51 & 93.37 & 91.97 & 88.92 \\
            SparseDet (ICCV'23) & 10.06 & 12.60 & 15.82 & 94.34 & 91.80 & 88.97 \\
            Calibrated Teacher (AAAI'23) & \underline{9.41} & \underline{12.10} & \underline{15.57} & \underline{95.27} & \underline{93.75} & \underline{91.51}  \\
            \cdashline{1-7}
            \rule{0pt}{10.0pt} 
            \bf SAMPD (Ours) & \bf 7.65 & \bf 9.03 & \bf 11.71 & \bf 95.38 & \bf94.39 & \bf92.08 \\
            \Xhline{3\arrayrulewidth}
            \end{tabular}
            }
\caption{Detection results (MR and AP$_{50}$) on the LLVIP dataset by varying removal percentages (30\%, 50\%, and 70\%) for sparsely annotated scenarios. \textbf{Bold}/\underline{underlined} fonts indicate the best/second-best results.}
\label{table:main_LLVIP}
\label{t3}
\end{table}

\begin{figure*}[t] 
  \centering
  \includegraphics[width=0.999\linewidth]{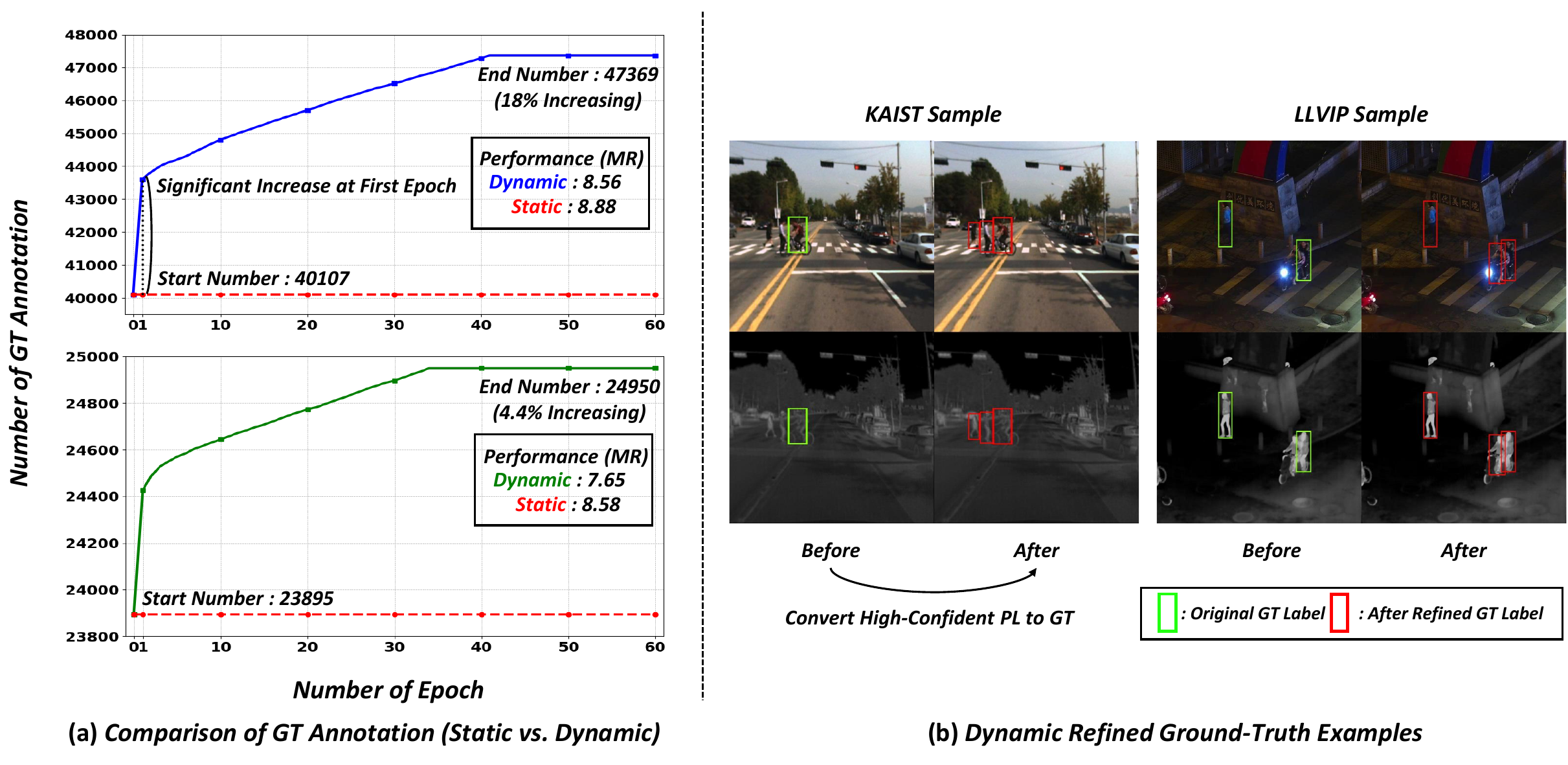}
    \caption{(a) Line graph of the changes in ground-truth annotations, and (b) visualization examples of the refined ground-truth. It shows that our APRA module (`Dynamic') effectively captures missed annotations in sparsely annotated environments.}
  \label{fig:static_dynamic}
\end{figure*}

\begin{table}[t!]
   \renewcommand{\tabcolsep}{4.2mm}
   \centering
   \begin{center}
      \resizebox{0.999\linewidth}{!}
      {
         \begin{tabular}{c c c ccc}
            \Xhline{3\arrayrulewidth}
             \rule{0pt}{10pt} \multirow{2}{*}[-0.3em]{\textbf{MPAW}} & \multirow{2}{*}[-0.3em]{\textbf{PPE}} & \multirow{2}{*}[-0.3em]{\bf APRA } & \multicolumn{3}{c}{\bf 30\%}\\ \cmidrule(lr){4-6}
            & & & \bf All & \bf Day & \bf Night\\\hline
            \rule{0pt}{9pt}
            - & - & - & 13.89 & 15.50 & 10.88  \\\cdashline{1-6}
            \rule{0pt}{9pt}\cmark & - & - & 10.14 & 11.18 & 7.82  \\
            \cmark & \cmark & - & 9.65 & 10.89 & 7.00 \\
            \cmark & - & \cmark & 9.23 & 10.76 &  6.26 \\
            \cmark & \cmark & \cmark & \bf 8.56 & \bf 10.55 & \bf 5.62 \\
            \Xhline{3\arrayrulewidth}
         \end{tabular}
      }
    \caption{Ablation studies of our multispectral pedestrian-aware adaptive weight (MPAW), positive pseudo-label enhancement (PPE), and adaptive pedestrian retrieval augmentation (APRA) on the KAIST dataset (30\% removal).}
    \label{table:ablation_study}
   \end{center}
\end{table}

\noindent\textbf{Evaluation Metric.}
Following \cite{kim2022towards, kim2021mlpd}, we measure the performance by adopting miss rate (MR), which is averaged over the range of false positives per image (FPPI), spanning from [$10^{-2}$, $10^0$]. The detection performance improves as the MR decreases. Following \cite{kim2021mlpd}, evaluations were conducted across three distinct settings: `All', `Day', and `Night'. For the LLVIP dataset, we also use Average Precision (AP) with an Intersection over Union (IoU) threshold of 0.5 (AP$_{50}$).

\subsection{Implementation Details}
We deploy the SAOD methodology leveraging an SSD \cite{liu2016ssd} structure combined with a VGG16 \cite{simonyan2015deep} backbone. We optimize our framework using Stochastic Gradient Descent (SGD) \cite{kiefer1952stochastic}, coordinating the process across two GTX 3090 GPUs and processing 6 images in each mini-batch. The augmentation include random horizontal flipping, color jittering, and cutout. Parameters are set with $m = 1$, $\tau = 0.1$ and $\lambda_1=\lambda_2=1$. We train our detector for 80 epochs with 0.0001 learning rate. All experimental procedures are performed utilizing the Pytorch framework \cite{paszke2017automatic}.

\subsection{Comparisons}
\noindent\textbf{Results on the KAIST Dataset.}
Table \ref{table:KAIST} shows the performance of our SAMPD with the state-of-the-art SAOD methods \cite{niitani2019sampling, zhang2020solving, wang2021comining, suri2023sparsedet, wang2023calibratedteachersparselyannotated} on the KAIST dataset. As removal percentages decrease from 70\% to 30\%, existing methods showed some improvement over the baseline (`Supervised'), which is trained only on sparsely annotated bounding boxes. However, SAMPD outperforms them across `All', `Day', and `Night' settings, with notable advantages at a 70\% removal rate. It highlights the effectiveness of our method in commonly encountered real-world scenarios with sparse or missing annotations. \\

\noindent\textbf{Results on the LLVIP Dataset.}
We also conduct experiments on the LLVIP dataset by varying removal percentages (30\%, 50\%, and 70\%). As shown in Table \ref{table:main_LLVIP}, Calibrated Teacher \cite{wang2023calibratedteachersparselyannotated} achieves the highest improvements among the existing methods. In contrast, our approach outperforms the Calibrated Teacher. By incorporating the proposed components during the training phase, our SAMPD effectively addresses sparsely annotated scenarios.

\begin{table}[t!]
     \renewcommand{\tabcolsep}{0.8mm}
   \centering
   \begin{center}
      \resizebox{0.999\linewidth}{!}
      {
         \begin{tabular}{c c c ccc}
            \Xhline{3\arrayrulewidth}
             \rule{0pt}{10pt} \multirow{2}{*}[-0.25em]{\bf \makecell{Augmentation \\ Methods}} & \multirow{2}{*}[-0.25em]{\textbf{MPAW}} & \multirow{2}{*}[-0.3em]{\bf PPE } & \multicolumn{3}{c}{\bf 30\%}\\ \cmidrule(lr){4-6}
            & & & \bf All & \bf Day & \bf Night\\\hline
            \rule{0pt}{8.5pt}
            Supervised  & - & - & 13.89           & 15.50           & 10.88 \\\cdashline{1-6}
            \rule{0pt}{9pt}
            Robust Teacher (CVIU'23) & \cmark & - & 11.25 & 12.85 & 8.09  \\
            \bf Ours (Static) & \cmark & - &  9.35 & 10.86 & 6.51 \\ \
            \bf Ours (Dynamic) & \cmark & - & \bf 9.23 & \bf  10.76 & \bf 6.26 \\\hline
            \rule{0pt}{9pt}
            Robust Teacher (CVIU'23) & \cmark & \cmark & 10.45 & 12.44 &  6.82 \\
            \bf Ours (Static) & \cmark & \cmark & 8.88 & 10.57 & 6.29 \\
            \bf Ours (Dynamic) & \cmark & \cmark & \bf 8.56 & \bf 10.55 & \bf 5.62 \\
            \Xhline{3\arrayrulewidth}
         \end{tabular}
      }
    \caption{Effect of pedestrian augmentation method. `Static' refers to our APRA method without a dynamic adding mechanism and `Dynamic' refers dynamic adding mechanism.}
   \label{table:Mixing}
   \end{center}
\end{table}

\subsection{Ablation Study}
We conduct the ablation study to explore the effect of the proposed modules, \textit{i.e.,} multispectral pedestrian-aware adaptive weight (MPAW) module, positive pseudo-label enhancement (PPE) module, and adaptive pedestrian retrieval augmentation (APRA) module. We conducted the ablation study with 30\% sparsely annotated scenario on the KAIST dataset.
As shown in Table \ref{table:ablation_study}, with the consideration of each module, the performances of our method are consistently improved. When all our modules are considered, we achieve the highest performance. In summary, our SAMPD effectively learns pedestrian representations in sparse annotation scenarios by: (1) reducing the impact of low-quality pseudo-labels with the multispectral pedestrian-aware adaptive weight $W_{AL}$ in MPAW module, (2) using PPE module to enhance the teacher model in generating higher-quality pseudo-labels, and (3) incorporating a wide range of pedestrian patches through APRA module to enrich pedestrian knowledge during training.

\subsection{Discussions}
\noindent\textbf{APRA Module (Static vs. Dynamic (Ours)).}
Our APRA module has a dynamic property that uses high-quality pseudo-labels as new ground-truths. Figure \ref{fig:static_dynamic} illustrates the changes over epochs during training on the KAIST and LLVIP datasets. As shown in Figure \ref{fig:static_dynamic}(a), the number of annotations increases progressively with each epoch. The dynamic mechanism of the APRA module learns various visual appearances of potentially missing pedestrians and shows improved performance compared to the static approach (\textit{i.e.,} without dynamic refining), with improvements such as 8.88 to 8.56 (KAIST dataset) and 8.58 to 7.65 (LLVIP dataset). In addition, Figure \ref{fig:static_dynamic}(b) demonstrates that the dynamic mechanism significantly reduces annotation sparsity, successfully filling in many previously missing annotations. \\

\noindent\textbf{Pedestrian Augmentation.}
To see the effectiveness of our APRA module, we compared it with the recent pedestrian augmentation approach, Robust Teacher \cite{LI2023103788}. As shown in Table \ref{table:Mixing}, while the Robust Teacher improved performance over the baseline (`Supervised'), our dynamic approach of APRA module outperformed it. In fact, since Robust Teacher does not account for scene-specific characteristics (\textit{e.g.,} scene brightness, patch sizes, etc.) leads to boundary bias and visual distortion from augmentation that do not match the lighting conditions of the scene. As a result, it performs worse compared to the static approach (w/o dynamic refinement). \\ 

\begin{table}[t!]
    \renewcommand{\tabcolsep}{1.4mm}
    \centering
    \resizebox{0.999\linewidth}{!}{
    \begin{tabular}{c c ccc}
        \Xhline{3\arrayrulewidth}
    \rule{0pt}{10pt}\bf Dataset & \bf $\mathcal{R}$ & \bf All & \bf Day & \bf Night\\ \hline
        \rule{0pt}{9.0pt}
            \bf \multirow{4}{*}{\bf KAIST}
                &  70\%  & 23.52 & 26.15 & 17.87 \\
                &  50\%  & 15.27 & 17.28 & 11.15 \\
                &  30\%  &  8.56 &  10.55 & \bf 5.62 \\\cdashline{2-5}
                \rule{0pt}{10.5pt}
                & 0\% &  \bf 7.58 &  \bf 7.95 & 6.95 \\\Xhline{3\arrayrulewidth}
            \end{tabular}
            \enskip
            \begin{tabular}{c c c}
        \Xhline{3\arrayrulewidth}
    \rule{0pt}{10pt}\bf Dataset & \bf $\mathcal{R}$ & \bf All \\ \hline
        \rule{0pt}{9.0pt}
            \bf \multirow{4}{*}{\bf LLVIP}
                &  70\%  & 11.71  \\
                &  50\%  & 9.73  \\
                &  30\%  & 7.65  \\\cdashline{2-3}
                \rule{0pt}{10.5pt}
                & 0\% & \bf 6.01\\
        \Xhline{3\arrayrulewidth}
    \end{tabular}
}
\caption{Comparison of performance (MR) between our SAMPD at different removal percentages ($\mathcal{R}$: 30\%, 50\%, 70\%) fully annotated baseline (0\% annotation removal scenario) on the KAIST and LLVIP datasets.}
\label{table:Fully}
\end{table}

\begin{table}[t!]
    \renewcommand{\tabcolsep}{0.5mm}
	\centering
        \resizebox{0.999\linewidth}{!}
		{
	\begin{tabular}{c ccc}
            \Xhline{3\arrayrulewidth}
            \rule{0pt}{10pt}\multirow{2}{*}{\bf Methods} & \multicolumn{3}{c}{\bf 0\% (Fully Labeled)} \\\cmidrule(lr){2-4} 
            & \bf All & \bf Day & \bf Night \\ \hline 
            \rule{0pt}{10.2pt} 
            Supervised & 7.58 &  7.95 & 6.95 \\\cdashline{1-4}
            \rule{0pt}{10.2pt} 
            Calibrated Teacher (AAAI'23) & 9.87 (-2.29) & 11.25 (-3.30)& 7.24 (-0.29) \\
            \bf SAMPD (Ours) & \bf 6.50 (+1.08) & \bf 6.85 (+1.10) & \bf 5.99 (+0.96)\\
            \Xhline{3\arrayrulewidth}
            \end{tabular}
            }
\caption{Detection results (MR) of our SAMPD and Calibrated Teacher in the fully annotated setting (0\% annotation removal scenario) on the KAIST dataset.}
\vspace{-0.2cm}
\label{table:supervised}
\end{table}


\noindent\textbf{SAMPD in Sparsely-/Fully-Annotated Scenarios.}
Table \ref{table:Fully} compares the performance our SAMPD at various removal percentages (30\%, 50\%, and 70\%) with the fully annotated baseline (0\% removal) on the KAIST and LLVIP datasets. Our method at 30\% removal percentage shows comparable performance to the fully annotated scenarios of the baseline. Interestingly, despite the 30\% removal percentage of annotations in the KAIST dataset, our method shows an improvement performance (1.33 MR) on `Night' setting.

Also, Table \ref{table:supervised} shows that applying our method in the 0\% removal scenario improves performance across all cases. This demonstrates that our SAMPD, by diversifying pedestrian sample knowledge through the proposed modules, is effective in both sparsely annotated environment and fully annotated environment, \textit{where potential incompleteness may exist}. \\


\noindent\textbf{Limitation.}
In SAMPD, while various results show the effectiveness of the APRA module, it currently offers only image-level guidance and samples pedestrians based on overall image brightness rather than individual characteristics. Thus, exploring a method to effectively guide at both the image-level and feature-level, considering individual pedestrian characteristics in addition to the entire image, can be a promising avenue for our future work.

\begin{figure}[t] 
  \centering
  \includegraphics[width=0.91\linewidth]{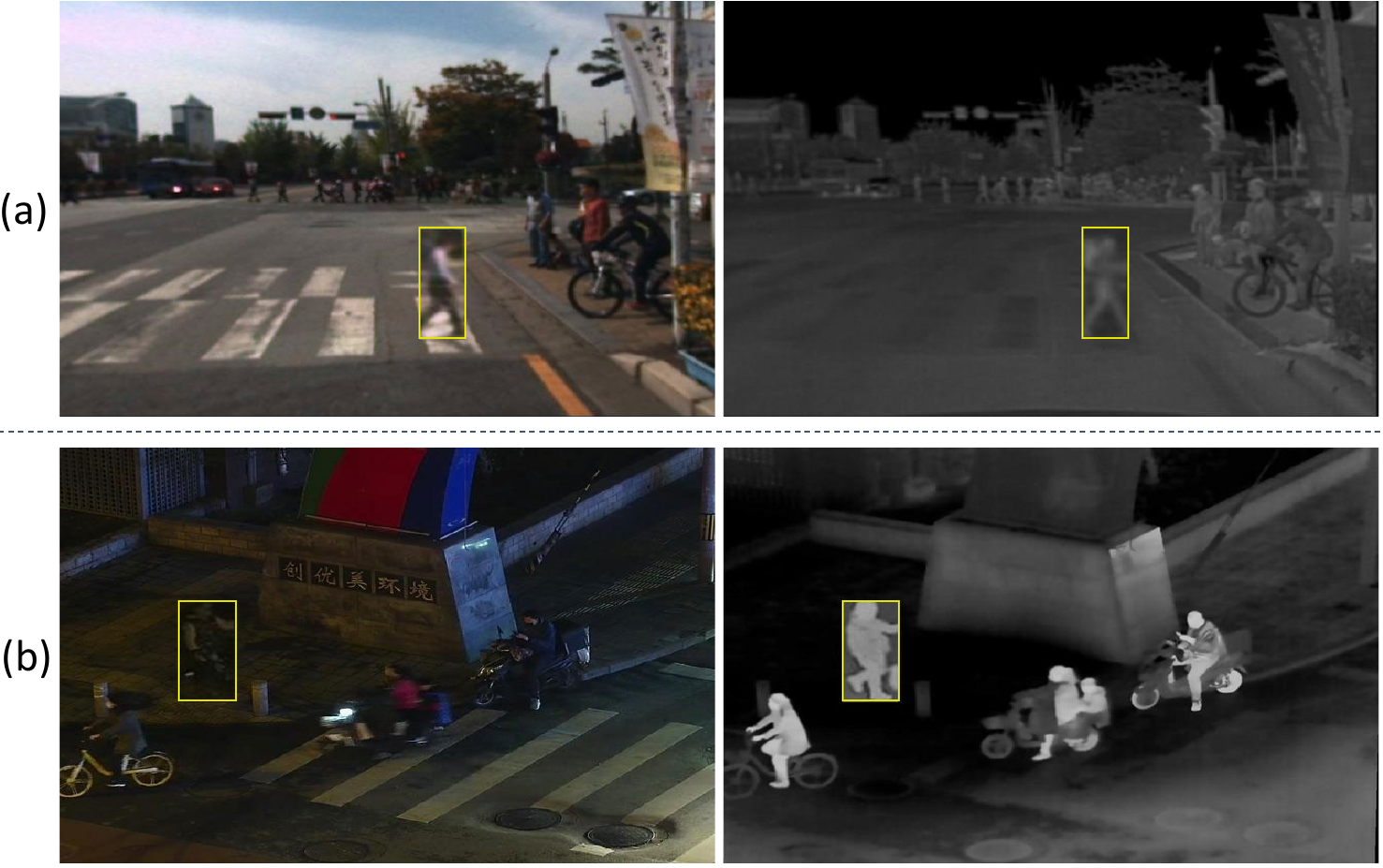}
  \caption{Visualization of APRA module on (a) KAIST dataset and (b) LLVIP dataset (left: visible, right: thermal). Yellow boxes are pedestrian patches from APRA module.}
  \label{fig:6}
  \vspace{-0.2cm}
\end{figure}

\subsection{Visualization Results}
\noindent\textbf{Effect of Adaptive Pedestrian Retrieval Augmentation.}
The APRA module enriches pedestrian information by attaching patches with similar brightness to the original image, ensuring smooth integration. Figure \ref{fig:6} shows examples of this augmentation of APRA module on the KAIST and LLVIP datasets. The APRA module integrates these patches without disrupting original image by placing them only in low-saliency regions, preserving visual appearance of pedestrians. More examples are in the supplementary document.

\section{Conclusion}
We present a new framework for multispectral pedestrian detection in sparsely annotated scenarios. Our method includes three key modules: MPAW module to increase the transfer weight of reliable pseudo-label knowledge, PPE module to guide the teacher model in generating better pseudo-labels, and APRA module to adaptively refine ground-truth annotations with pedestrian patches. Experimental results demonstrate the effectiveness of our approach in handling sparse annotations. We believe our method provides valuable insights for various sparsely annotated scenarios.

\section{Acknowledgements}
This work was supported by the NRF grant funded by the Korea government (MSIT) (No. RS-2023-00252391), and by IITP grant funded by the Korea government (MSIT) (No. RS-2022-00155911: Artificial Intelligence Convergence Innovation Human Resources Development (Kyung Hee University), IITP-2022-II220078: Explainable Logical Reasoning for Medical Knowledge Generation, No. RS-2024-00509257: Global AI Frontier Lab), and by the MSIT (Ministry of Science and ICT), Korea, under the National Program for Excellence in SW (2023-0-00042) supervised by the IITP in 2025, and conducted by CARAI grant funded by DAPA and ADD (UD230017TD).

\bibliography{aaai25}

\begin{thebibliography}{36}
\providecommand{\natexlab}[1]{#1}

\bibitem[{Chen et~al.(2023)Chen, Xie, Nie, Cao, Shao, and Pang}]{chen2023attentive}
Chen, N.; Xie, J.; Nie, J.; Cao, J.; Shao, Z.; and Pang, Y. 2023.
\newblock Attentive alignment network for multispectral pedestrian detection.
\newblock In \emph{Proceedings of the 31st ACM international conference on multimedia}, 3787--3795.

\bibitem[{Chen et~al.(2022)Chen, Shi, Ye, Mertz, Ramanan, and Kong}]{chen2022multimodal}
Chen, Y.-T.; Shi, J.; Ye, Z.; Mertz, C.; Ramanan, D.; and Kong, S. 2022.
\newblock Multimodal object detection via probabilistic ensembling.
\newblock In \emph{European Conference on Computer Vision}, 139--158. Springer.

\bibitem[{Dasgupta et~al.(2022)Dasgupta, Das, Das, Bhattacharya, and Yogamani}]{9706418}
Dasgupta, K.; Das, A.; Das, S.; Bhattacharya, U.; and Yogamani, S. 2022.
\newblock Spatio-Contextual Deep Network-Based Multimodal Pedestrian Detection for Autonomous Driving.
\newblock \emph{IEEE Transactions on Intelligent Transportation Systems}, 23(9): 15940--15950.

\bibitem[{Guan et~al.(2019)Guan, Cao, Yang, Cao, and Yang}]{guan2019fusion}
Guan, D.; Cao, Y.; Yang, J.; Cao, Y.; and Yang, M.~Y. 2019.
\newblock Fusion of multispectral data through illumination-aware deep neural networks for pedestrian detection.
\newblock \emph{Information Fusion}, 50: 148--157.

\bibitem[{Hu, Zhang, and Weng(2023)}]{10222560}
Hu, Y.; Zhang, N.; and Weng, L. 2023.
\newblock Retrieve the Visible Feature to Improve Thermal Pedestrian Detection Using Discrepancy Preserving Memory Network.
\newblock In \emph{2023 IEEE International Conference on Image Processing (ICIP)}, 1125--1129.

\bibitem[{Hwang et~al.(2015)Hwang, Park, Kim, Choi, and So~Kweon}]{hwang2015multispectral}
Hwang, S.; Park, J.; Kim, N.; Choi, Y.; and So~Kweon, I. 2015.
\newblock Multispectral pedestrian detection: Benchmark dataset and baseline.
\newblock In \emph{Proceedings of the IEEE conference on computer vision and pattern recognition}, 1037--1045.

\bibitem[{Jia et~al.(2021)Jia, Zhu, Li, Tang, and Zhou}]{jia2021llvip}
Jia, X.; Zhu, C.; Li, M.; Tang, W.; and Zhou, W. 2021.
\newblock LLVIP: A visible-infrared paired dataset for low-light vision.
\newblock In \emph{Proceedings of the IEEE/CVF international conference on computer vision}, 3496--3504.

\bibitem[{Kiefer and Wolfowitz(1952)}]{kiefer1952stochastic}
Kiefer, J.; and Wolfowitz, J. 1952.
\newblock Stochastic estimation of the maximum of a regression function.
\newblock \emph{The Annals of Mathematical Statistics}, 462--466.

\bibitem[{Kim et~al.(2021)Kim, Kim, Kim, Kim, and Choi}]{kim2021mlpd}
Kim, J.; Kim, H.; Kim, T.; Kim, N.; and Choi, Y. 2021.
\newblock MLPD: Multi-label pedestrian detector in multispectral domain.
\newblock \emph{IEEE Robotics and Automation Letters}, 6(4): 7846--7853.

\bibitem[{Kim, Park, and Ro(2021{\natexlab{a}})}]{kim2021robust}
Kim, J.~U.; Park, S.; and Ro, Y.~M. 2021{\natexlab{a}}.
\newblock Robust small-scale pedestrian detection with cued recall via memory learning.
\newblock In \emph{Proceedings of the IEEE/CVF International Conference on Computer Vision}, 3050--3059.

\bibitem[{Kim, Park, and Ro(2021{\natexlab{b}})}]{kim2021uncertainty}
Kim, J.~U.; Park, S.; and Ro, Y.~M. 2021{\natexlab{b}}.
\newblock Uncertainty-guided cross-modal learning for robust multispectral pedestrian detection.
\newblock \emph{IEEE Transactions on Circuits and Systems for Video Technology}, 32(3): 1510--1523.

\bibitem[{Kim, Park, and Ro(2022)}]{kim2022towards}
Kim, J.~U.; Park, S.; and Ro, Y.~M. 2022.
\newblock Towards versatile pedestrian detector with multisensory-matching and multispectral recalling memory.
\newblock In \emph{Proceedings of the AAAI Conference on Artificial Intelligence}, volume 36(1), 1157--1165.

\bibitem[{Kim and Ro(2023)}]{kim2023similarity}
Kim, J.~U.; and Ro, Y.~M. 2023.
\newblock Similarity Relation Preserving Cross-Modal Learning for Multispectral Pedestrian Detection Against Adversarial Attacks.
\newblock In \emph{ICASSP 2023-2023 IEEE International Conference on Acoustics, Speech and Signal Processing (ICASSP)}, 1--5. IEEE.

\bibitem[{Kim et~al.(2024)Kim, Shin, Yu, Kim, and Ro}]{kim2024causal}
Kim, T.; Shin, S.; Yu, Y.; Kim, H.~G.; and Ro, Y.~M. 2024.
\newblock Causal Mode Multiplexer: A Novel Framework for Unbiased Multispectral Pedestrian Detection.
\newblock In \emph{Proceedings of the IEEE/CVF Conference on Computer Vision and Pattern Recognition}, 26784--26793.

\bibitem[{Lewis et~al.(2021)Lewis, Perez, Piktus, Petroni, Karpukhin, Goyal, Küttler, Lewis, tau Yih, Rocktäschel, Riedel, and Kiela}]{lewis2021retrievalaugmentedgenerationknowledgeintensivenlp}
Lewis, P.; Perez, E.; Piktus, A.; Petroni, F.; Karpukhin, V.; Goyal, N.; Küttler, H.; Lewis, M.; tau Yih, W.; Rocktäschel, T.; Riedel, S.; and Kiela, D. 2021.
\newblock Retrieval-Augmented Generation for Knowledge-Intensive NLP Tasks.
\newblock arXiv:2005.11401.

\bibitem[{Li et~al.(2019)Li, Song, Tong, and Tang}]{li2019illumination}
Li, C.; Song, D.; Tong, R.; and Tang, M. 2019.
\newblock Illumination-aware faster R-CNN for robust multispectral pedestrian detection.
\newblock \emph{Pattern Recognition}, 85: 161--171.

\bibitem[{Li et~al.(2023)Li, Liu, Shen, Sun, and Tan}]{LI2023103788}
Li, S.; Liu, J.; Shen, W.; Sun, J.; and Tan, C. 2023.
\newblock Robust Teacher: Self-correcting pseudo-label-guided semi-supervised learning for object detection.
\newblock \emph{Computer Vision and Image Understanding}, 235: 103788.

\bibitem[{Liu et~al.(2016)Liu, Anguelov, Erhan, Szegedy, Reed, Fu, and Berg}]{liu2016ssd}
Liu, W.; Anguelov, D.; Erhan, D.; Szegedy, C.; Reed, S.; Fu, C.-Y.; and Berg, A.~C. 2016.
\newblock Ssd: Single shot multibox detector.
\newblock In \emph{Computer Vision--ECCV 2016: 14th European Conference, Amsterdam, The Netherlands, October 11--14, 2016, Proceedings, Part I 14}, 21--37. Springer.

\bibitem[{Liu et~al.(2024)Liu, Hu, Zhao, Huang, and Zhang}]{liu2024region}
Liu, Y.; Hu, C.; Zhao, B.; Huang, Y.; and Zhang, X. 2024.
\newblock Region-Based Illumination-Temperature Awareness and Cross-Modality Enhancement for Multispectral Pedestrian Detection.
\newblock \emph{IEEE Transactions on Intelligent Vehicles}.

\bibitem[{Niitani et~al.(2019)Niitani, Akiba, Kerola, Ogawa, Sano, and Suzuki}]{niitani2019sampling}
Niitani, Y.; Akiba, T.; Kerola, T.; Ogawa, T.; Sano, S.; and Suzuki, S. 2019.
\newblock Sampling Techniques for Large-Scale Object Detection from Sparsely Annotated Objects.
\newblock arXiv:1811.10862.

\bibitem[{Park et~al.(2022)Park, Choi, Kim, and Ro}]{park2022robust}
Park, S.; Choi, D.~H.; Kim, J.~U.; and Ro, Y.~M. 2022.
\newblock Robust thermal infrared pedestrian detection by associating visible pedestrian knowledge.
\newblock In \emph{ICASSP 2022-2022 IEEE International Conference on Acoustics, Speech and Signal Processing (ICASSP)}, 4468--4472. IEEE.

\bibitem[{Paszke et~al.(2017)Paszke, Gross, Chintala, Chanan, Yang, DeVito, Lin, Desmaison, Antiga, and Lerer}]{paszke2017automatic}
Paszke, A.; Gross, S.; Chintala, S.; Chanan, G.; Yang, E.; DeVito, Z.; Lin, Z.; Desmaison, A.; Antiga, L.; and Lerer, A. 2017.
\newblock Automatic differentiation in PyTorch.
\newblock \emph{NeurIPS-W}.

\bibitem[{Ramos, Elliott, and Martins(2023)}]{ramos2023retrievalaugmentedimagecaptioning}
Ramos, R.; Elliott, D.; and Martins, B. 2023.
\newblock Retrieval-augmented Image Captioning.
\newblock arXiv:2302.08268.

\bibitem[{Sarto et~al.(2022)Sarto, Cornia, Baraldi, and Cucchiara}]{sarto2022retrievalaugmentedtransformerimagecaptioning}
Sarto, S.; Cornia, M.; Baraldi, L.; and Cucchiara, R. 2022.
\newblock Retrieval-Augmented Transformer for Image Captioning.
\newblock arXiv:2207.13162.

\bibitem[{Simonyan and Zisserman(2015)}]{simonyan2015deep}
Simonyan, K.; and Zisserman, A. 2015.
\newblock Very Deep Convolutional Networks for Large-Scale Image Recognition.
\newblock arXiv:1409.1556.

\bibitem[{Suri et~al.(2023)Suri, Rambhatla, Chellappa, and Shrivastava}]{suri2023sparsedet}
Suri, S.; Rambhatla, S.~S.; Chellappa, R.; and Shrivastava, A. 2023.
\newblock SparseDet: Improving Sparsely Annotated Object Detection with Pseudo-positive Mining.
\newblock arXiv:2201.04620.

\bibitem[{Wang et~al.(2023{\natexlab{a}})Wang, Liu, Zhang, Zhang, Zhang, Gan, Wang, Wang, and Wang}]{wang2023calibratedteachersparselyannotated}
Wang, H.; Liu, L.; Zhang, B.; Zhang, J.; Zhang, W.; Gan, Z.; Wang, Y.; Wang, C.; and Wang, H. 2023{\natexlab{a}}.
\newblock Calibrated Teacher for Sparsely Annotated Object Detection.
\newblock arXiv:2303.07582.

\bibitem[{Wang et~al.(2021)Wang, Yang, Cao, and Zhang}]{wang2021comining}
Wang, T.; Yang, T.; Cao, J.; and Zhang, X. 2021.
\newblock Co-mining: Self-Supervised Learning for Sparsely Annotated Object Detection.
\newblock arXiv:2012.01950.

\bibitem[{Wang et~al.(2023{\natexlab{b}})Wang, Yang, Zhang, Li, Feng, Fang, Lyu, Chen, and Zhang}]{wang2023consistentteacher}
Wang, X.; Yang, X.; Zhang, S.; Li, Y.; Feng, L.; Fang, S.; Lyu, C.; Chen, K.; and Zhang, W. 2023{\natexlab{b}}.
\newblock Consistent-Teacher: Towards Reducing Inconsistent Pseudo-targets in Semi-supervised Object Detection.
\newblock arXiv:2209.01589.

\bibitem[{Xie et~al.(2022)Xie, Anwer, Cholakkal, Nie, Cao, Laaksonen, and Khan}]{xie2022learning}
Xie, J.; Anwer, R.~M.; Cholakkal, H.; Nie, J.; Cao, J.; Laaksonen, J.; and Khan, F.~S. 2022.
\newblock Learning a dynamic cross-modal network for multispectral pedestrian detection.
\newblock In \emph{Proceedings of the 30th ACM International Conference on Multimedia}, 4043--4052.

\bibitem[{Xie et~al.(2024)Xie, Cheng, Zhong, Zhou, Markham, and Trigoni}]{xie2024beyond}
Xie, Q.; Cheng, T.-Y.; Zhong, J.-X.; Zhou, K.; Markham, A.; and Trigoni, N. 2024.
\newblock Beyond Fusion: Modality Hallucination-based Multispectral Fusion for Pedestrian Detection.
\newblock In \emph{Proceedings of the IEEE/CVF Winter Conference on Applications of Computer Vision}, 655--664.

\bibitem[{Xu et~al.(2023)Xu, Zhan, Zhu, Jiang, Chen, and Guo}]{e25071022}
Xu, X.; Zhan, W.; Zhu, D.; Jiang, Y.; Chen, Y.; and Guo, J. 2023.
\newblock Contour Information-Guided Multi-Scale Feature Detection Method for Visible-Infrared Pedestrian Detection.
\newblock \emph{Entropy}, 25(7).

\bibitem[{Zhang et~al.(2020)Zhang, Chen, Shen, Hao, Zhu, and Savvides}]{zhang2020solving}
Zhang, H.; Chen, F.; Shen, Z.; Hao, Q.; Zhu, C.; and Savvides, M. 2020.
\newblock Solving Missing-Annotation Object Detection with Background Recalibration Loss.
\newblock arXiv:2002.05274.

\bibitem[{Zhang et~al.(2019)Zhang, Zhu, Chen, Yang, Lei, and Liu}]{zhang2019weakly}
Zhang, L.; Zhu, X.; Chen, X.; Yang, X.; Lei, Z.; and Liu, Z. 2019.
\newblock Weakly aligned cross-modal learning for multispectral pedestrian detection.
\newblock In \emph{Proceedings of the IEEE/CVF international conference on computer vision}, 5127--5137.

\bibitem[{Zhou, Chen, and Cao(2020)}]{zhou2020improving}
Zhou, K.; Chen, L.; and Cao, X. 2020.
\newblock Improving multispectral pedestrian detection by addressing modality imbalance problems.
\newblock In \emph{Computer Vision--ECCV 2020: 16th European Conference, Glasgow, UK, August 23--28, 2020, Proceedings, Part XVIII 16}, 787--803. Springer.

\bibitem[{Zhu et~al.(2023)Zhu, Wu, Wang, He, Liu, and Pan}]{zhu2023dpacfuse}
Zhu, H.; Wu, H.; Wang, X.; He, D.; Liu, Z.; and Pan, X. 2023.
\newblock DPACFuse: Dual-Branch Progressive Learning for Infrared and Visible Image Fusion with Complementary Self-Attention and Convolution.
\newblock \emph{Sensors}, 23(16): 7205.

\end{thebibliography}
\clearpage

%

\def\maketitlesupplementary
   {
   \newpage
       \twocolumn[
        \centering
        \Large
        \textbf{Multispectral Pedestrian Detection with Sparsely Annotated Label\\--\textit{Supplementary Material}--}\\
        \vspace{1.5em}
       ] 
   }

\title{Multispectral Pedestrian Detection with Sparsely Annotated Label\\ --\textit {Supplementary Material} --}

\setcounter{secnumdepth}{0} 

\setcounter{table}{0}
\setcounter{figure}{0}

\maketitlesupplementary

\begin{algorithm}
\caption{Sparsely Annotated Environment Generation}
\footnotesize
\begin{algorithmic}
\Procedure{GenerateSA}{$N$, $paths$}
    \State $n \gets 0$
    \While{$n < N$}
        \For{$(p_V, p_T) \in paths$}
            \State $A_V \gets \Call{ReadFile}{p_V}$
            \State $A_T \gets \Call{ReadFile}{p_T}$
            \If{$\Call{Length}{A_V} > 1$}
                \State $areas_V \gets \Call{CalcAreas}{A_V}$
                \State $P_V \gets \Call{CalcProb}{areas_V}$
                \State $i \gets \Call{ArgMax}{P_V}$
                \State \Call{Delete}{$A_V$, $i$}
                \State $n \gets n + 1$
            \EndIf
            \If{$\Call{Length}{A_T} > 1$}
                \State $areas_T \gets \Call{CalcAreas}{A_T}$
                \State $P_T \gets \Call{CalcProb}{areas_T}$
                \State $i \gets \Call{ArgMax}{P_T}$
                \State \Call{Delete}{$A_T$, $i$}
                \State $n \gets n + 1$
            \EndIf
            \State \Call{WriteFile}{$p_V$, $A_V$}
            \State \Call{WriteFile}{$p_T$, $A_T$}
        \EndFor
    \EndWhile
\EndProcedure
\end{algorithmic}
\label{algorithm:reduce}
\end{algorithm}

This manuscript provides the additional results of the proposed method. Section \ref{sec:1} to \ref{sec:4} explains the details of our methods and experimental results to show the effectiveness of our methods, and Section \ref{sec:5} shows the additional visualization results.
Please note that Table PXX indicates the reference and the table number in the main paper.

\DeclarePairedDelimiter{\CalcProb*}{P}{}
\section{Sparsely Annotated Environment Generation}
\label{sec:1}

We describe the process of transforming a fully annotated dataset into a sparsely annotated environment, essential for validating our Sparsely Annotated Multispectral Pedestrian Detection (SAMPD) framework. The pseudo-code for this algorithm is outlined in Algorithm 1. `$V$' represents the visible component, `$T$' represents the thermal component, `$N$' represents the total number of annotations to delete, `$p$' represents the file path, `$A$' represents the annotations, and `$P$' denotes a list of probability values. The ``CalcProb'' function takes ``areas'' as input, where ``areas'' represents the sizes of the annotated boxes. It returns a list of probabilities, with smaller area values being assigned higher probabilities than larger ones. To simulate sparse annotations, we selectively removed 30\%, 50\%, and 70\% of the existing annotations. In Algorithm 1, begins by counting the total number of annotations in the dataset. Based on the desired sparsity percentage, we calculate the exact number of annotations to be deleted. Subsequently, we accessed the annotation files for both visible and thermal modalities, deleted the specified number of annotations, and saved the updated files. This process was repeated until the target reduction in annotations (sparsely annotation) was achieved.

\begin{table}[t]
    \renewcommand{\tabcolsep}{2.5mm}
	\centering
        \resizebox{0.999\linewidth}{!}
		{
	\begin{tabular}{c c c}
            \Xhline{3\arrayrulewidth}
            \rule{0pt}{9.0pt} 
            \multirow{2}{*}{\bf Methods} & \multicolumn{2}{c}{\bf 0\% (Fully Labeled)} \\\cmidrule(lr){2-3} 
            & \bf MR & \bf AP$_{50}$ \\ \hline 
            \rule{0pt}{9.0pt} 
            Supervised & 6.01 & 96.22 \\ \cdashline{1-3} 
            \rule{0pt}{9.0pt} 
            Calibrated Teacher (AAAI'23) & 6.81 (-0.80) & 95.80 (-0.42) \\
            \bf SAMPD (Ours) & \bf 5.70 (+0.31) & \bf 96.33 (+0.11) \\
            \Xhline{3\arrayrulewidth}
            \end{tabular}
            }
    \caption{Detection results (MR) of our SAMPD and the latest SAOD method,  Calibrated Teacher, in the fully annotated setting (0\% annotation drop scenario) on LLVIP dataset.}
\label{table:fully_LLVIP}
\end{table}


\section{Fully-Annotated Scenarios in LLVIP Dataset}
\label{sec:2}

Similar to Table P7 in the main paper, we conducted experiments to compare detection results under the fully annotated setting on the LLVIP dataset. We evaluated our method against the baseline (Supervised') and the latest SAOD approach, Calibrated Teacher \cite{wang2023calibratedteachersparselyannotated}. The results are shown in Table \ref{table:fully_LLVIP}. Even in the fully annotated setting (\textit{i.e.,} 0\% drop scenario), our method enhances the performance on the LLVIP dataset. Specifically, the baseline (`Supervised') shows 6.01 MR and 96.22 AP$_{50}$ and our method shows 5.70 MR and 96.33 AP$_{50}$, an increase of 0.31 MR and 0.11 AP$_{50}$. Compared to the 30\% annotation removal presented in Table P2 of the main paper, where the performance was 9.41 MR and 95.27 AP$_{50}$, the Calibrated Teacher shows improved performance with 6.81 MR and 95.80 AP$_{50}$. In contrast, our SAMPD further enhanced the performance. It demonstrates the effectiveness of our approach, which employs the positive pseudo-label enhancement (PPE) module to ensure the generation of high-quality pseudo-labels and the adaptive pedestrian retrieval augmentation (APRA) module to enrich pedestrian sample knowledge, across both sparsely and fully annotated environments.


\begin{table}[t!]
    \renewcommand{\tabcolsep}{2.8mm}
    \centering
    \resizebox{0.999\linewidth}{!}{
        \begin{tabular}{c c ccc}
            \Xhline{3\arrayrulewidth}
            \multirow{2}{*}{\textbf{Backbone}} & \multirow{2}{*}{\makecell{\textbf{Proposed}\\\textbf{Method}}} & \multicolumn{3}{c}{\bf 30\%} \\ \cmidrule(lr){3-5}
            & & \bf All &\bf Day & \bf Night  \\\hline
            \multirow{2}{*}{ResNet-50} & \ding{55} & 16.69 & 20.28 & 10.07 \\ 
            & \ding{51} & \textbf{13.17} & \textbf{16.35} & \textbf{7.22} \\ \cmidrule{1-5}
            \multirow{2}{*}{ResNet-101} & \ding{55} & 16.39 & 19.14 & 11.55 \\
            & \ding{51} & \textbf{13.07} & \textbf{15.89} & \textbf{8.47} \\ 
            \Xhline{3\arrayrulewidth}
        \end{tabular}
    }
    \caption{Comparison of different backbones with and without the proposed method.}
    \label{tab:comparison}
\end{table}

\begin{table}[t!]
     \renewcommand{\tabcolsep}{3.8mm}
   \centering
   \begin{center}
      \resizebox{0.999\linewidth}{!}
      {
         \begin{tabular}{c c ccc}
            \Xhline{3\arrayrulewidth}
            \multicolumn{2}{c}{$w^k$} & \multicolumn{3}{c}{\bf 30\%}\\\cmidrule(lr){1-2} \cmidrule(lr){3-5}
            \bf Single & \bf Fusion&\bf All & \bf Day & \bf Night\\\hline
            \rule{0pt}{9pt}
            \ding{51} & \ding{55} & 9.69 & 11.39 & 6.25  \\
             \ding{55} & \ding{51} &  9.31 & 10.91 &  6.30 \\ \
             \ding{51} & \ding{51} & \bf 8.56 & \bf  10.55 & \bf 5.62 \\\Xhline{3\arrayrulewidth}
         \end{tabular}
      }
    \caption{Effect of pedestrian MPAW module on the KAIST dataset (30\% removal).`Single' refers that $w^k$ apply to individual visible and thermal modalities detection loss, respectively, and `Fusion' refers that $w^k$ apply to fusion modality detection loss. }
   \label{table:single_fusion}
   \end{center}
\end{table}

\begin{table}[t!]
    \renewcommand{\tabcolsep}{0.8mm}
	\centering
        \resizebox{0.999\linewidth}{!}
		{
	\begin{tabular}{c ccc ccc}
            \Xhline{3\arrayrulewidth}
            \rule{0pt}{10pt} \multirow{2}{*}{\bf Methods} & \multicolumn{3}{c}{\bf Visible} & \multicolumn{3}{c}{\bf Thermal} \\ 
            \cmidrule(lr){2-4} \cmidrule(lr){5-7}
            &\bf  All         &\bf  Day         & \bf Night       &\bf  All         &\bf  Day         & \bf Night
            
            \\ \hline
            \rule{0pt}{10.0pt}
            Supervised & 28.38 & 22.65 & 40.56 & 20.64 & 26.19 & 9.01 \\ \cdashline{1-7} \rule{0pt}{10.2pt} 
            Calibrated Teacher (AAAI'23) & 25.91 & 21.73 & 35.80 & 17.23 & 21.35 & 9.04  \\

            \bf SAMPD (Ours) & \bf 23.56 & \bf 18.90 & \bf 35.57 & \bf16.34 & \bf19.88 & \bf8.14 \\
            \Xhline{3\arrayrulewidth}
            \end{tabular}
            }
        \caption{Detection results (MR) on the KAIST dataset for single modality (visible and thermal) with 30\% sparsely annotated scenarios. We compare our method with the state-of-the-art Calibrated Teacher.}
\label{table:single}
\label{t3}
\end{table}

\section{Generalization Ability of our SAMPD}
We conduct additional experiments to demonstrate the generalizability of our proposed framework. Specifically, we replaced the VGG16 backbone in the existing VGG16-SSD detector framework with ResNet-50 and ResNet-101. The results of these experiments are presented in Table \ref{tab:comparison}. As shown in the table, our method continues to perform effectively even when the backbone is changed, with improvements observed in the results. This indicates that our approach successfully generates pseudo-labels that enhance learning, particularly in scenarios with sparse labels. These results demonstrate that our method is effective not only within the original VGG16-SSD framework but is also adaptable to other backbone architectures.

\section{Effect of the Single and Fusion Modalities $w^k$}
We conducted ablation studies on the MPAW module to analyze the effectiveness of $w^k$ for each modality. As shown in Table \ref{table:single_fusion}, when considering only the weight of single or fusion modalities, the results were suboptimal (9.69 and 9.31 MR, respectively). However, when the weights of both single and fusion modalities were considered together, the performance improved significantly, achieving the best result of 8.56 MR. These findings indicate that our MPAW module effectively adjusts the knowledge from both single and fusion modalities.

\section{Utilization in Single Modality}
To demonstrate the effectiveness of our SAMPD in a single modality, we compared it with the Calibrated Teacher \cite{wang2023calibratedteachersparselyannotated}, which exhibits the highest performance performance among existing methods. As shown in Table \ref{table:single}, our method still outperforms the Calibrated Teacher across various single modalities (visible and thermal), indicating that SAMPD can further extend to single modal scenarios.

\begin{table}[t]
    \renewcommand{\tabcolsep}{4.3mm}
	\centering
        \vspace{-0.2cm}
        \resizebox{0.999\linewidth}{!}
		{
	\begin{tabular}{c ccc}
            \Xhline{3\arrayrulewidth}
            \multirow{2}{*}{\bf Methods} & \multirow{2}{*}{\bf All} & \multirow{2}{*}{\bf Day} & \multirow{2}{*}{\bf Night} \\ \\ \hline 
            \rule{0pt}{10.2pt} 
            Random Removal & 13.04 & 14.02 & 10.74 \\
            \bf Ours & 13.89 & 15.50 & 10.88 \\
            \Xhline{3\arrayrulewidth}
            \end{tabular}
            }
\caption{Comparison of performance (MR) between random removal and our removal approach (higher removal probability for bounding-box annotations with smaller widths) on KAIST dataset with a 30\% removal percentage. The results indicate that our approach is more realistic scenario and more challenging than random removal approach.}
\label{table:scenario}
\end{table}

\section{Sparsely Annotated Scenarios \\(Random Removal vs. Ours)}
\begin{figure*}[t] 
  \centering
  \includegraphics[width=\linewidth]{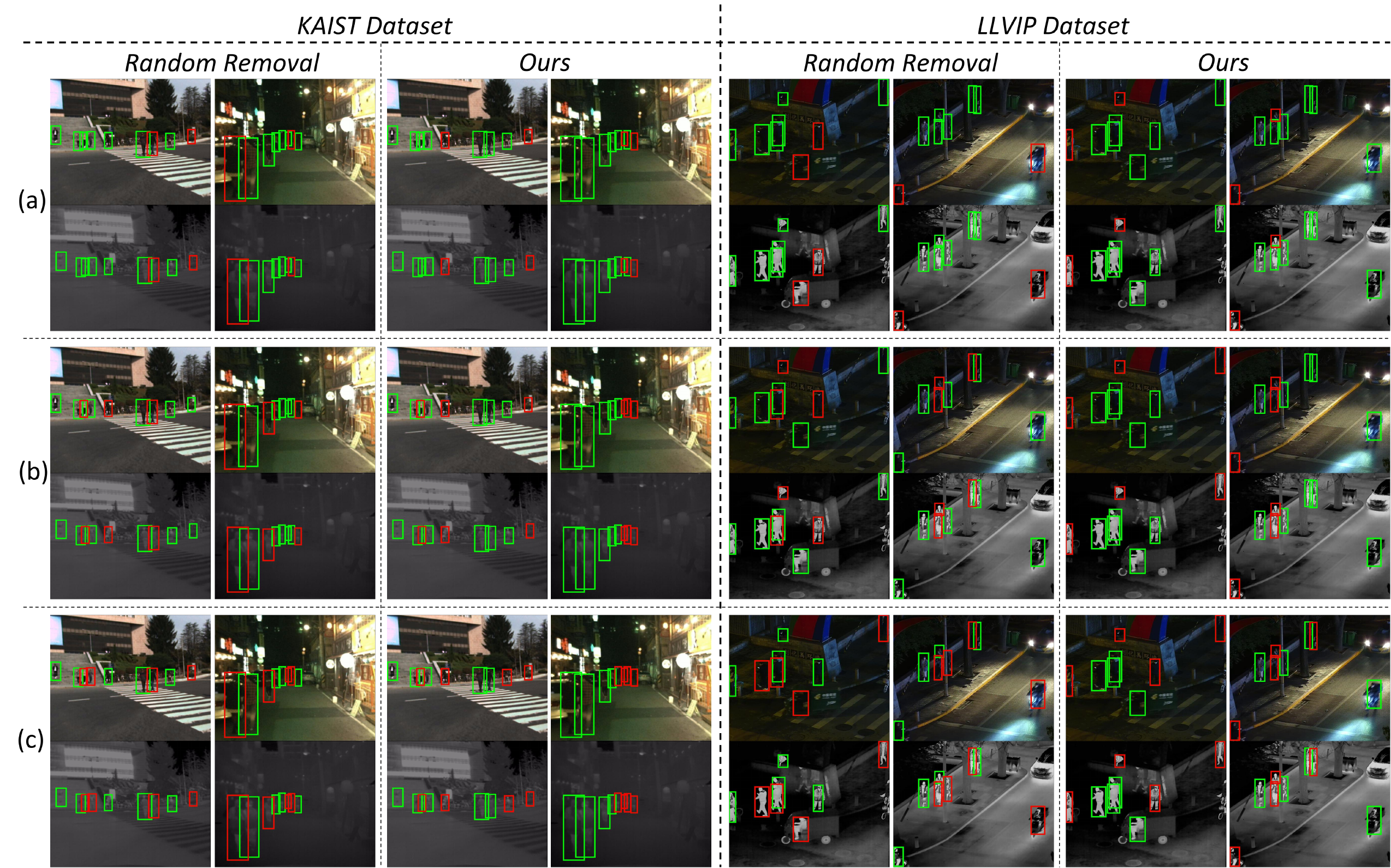}
  \caption{Visualization results of sparse annotation scenarios on KAIST dataset and LLVIP dataset (Green: ground-truth bounding-box and red: removed bounding-box). Each image pair indicates visible and thermal images, respectively. (a), (b), and (c) represent the removal percentages of 30\%, 50\%, and 70\%, respectively.}
  \label{fig:scenario_kaist}
\end{figure*}

We consider different scenarios of sparse annotations that might result from human error, focusing on two main strategies: (1) random removal of boxes or (2) targeted removal of smaller boxes (ours). We decide to focus on the scenario where smaller boxes are more likely to be omitted, based on the assumption that human error more frequently occurs through the misrecognition of smaller objects. Figure \ref{fig:scenario_kaist} provides a visual comparison of these two scenarios, illustrating that the preferential omission of smaller boxes is the scenario most reflective of real-world conditions. Table \ref{table:scenario} shows that the technique focusing on removing smaller boxes results in a greater decline in performance compared to the method that removes boxes randomly. This suggests that, similar to human perception, the model finds it more difficult to discern information about small objects, significantly hampering its learning process. Our removal method specifically targets the annotations for pedestrians with smaller box sizes, which are more likely to be missed by humans. In contrast, the random removal method often eliminates annotations for larger pedestrians, which is less common in real-life situations. The visualization results confirm that our removal method mimics human behavior, often overlooking small objects, thereby reflecting the realistic challenges encountered in real-world scenarios.\\

\begin{figure*}[t] 
  
  \centering
  \includegraphics[width=\linewidth]{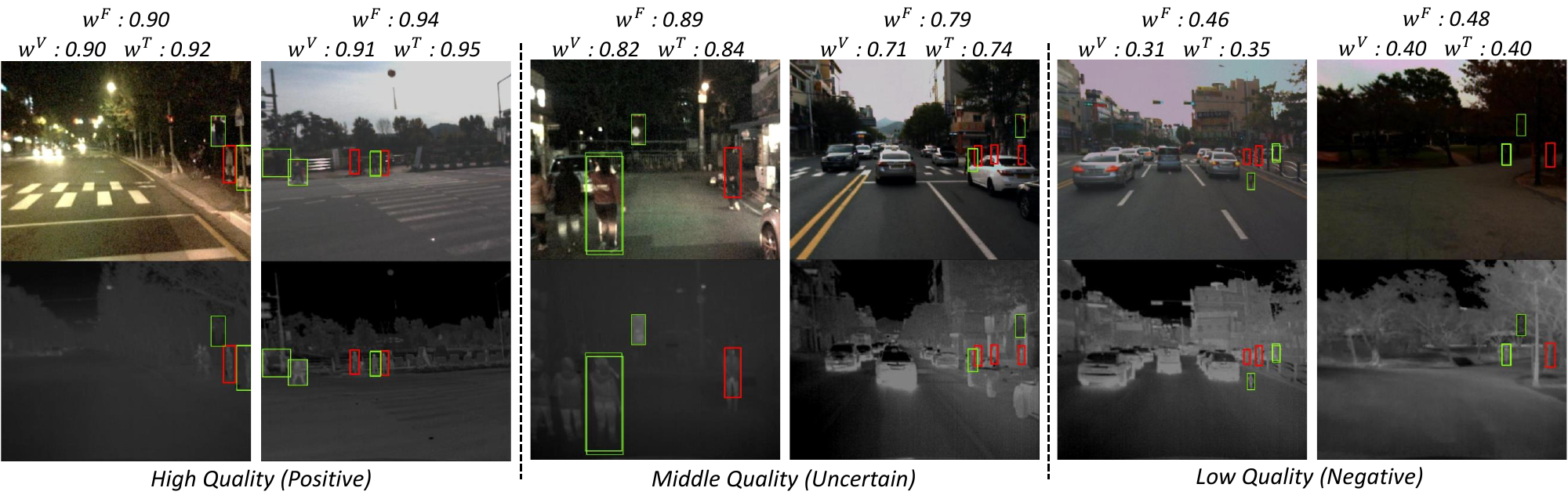}
  \vspace{-0.5cm}
  \caption{Detection results of the MPAW module on KAIST dataset. The vertical image pair indicates visible and thermal images. These image pairs are used for the training phase. Green bounding-boxes denote the ground-truth and red bounding-boxes indicate the predicted bounding-boxes. The `high quality' means the pseudo-label correctly captures the pedestrians, and the `low quality' means the pseudo-label incorrectly captures the pedestrians. In addition, `middle quality' means the pseudo-label captures the pedestrians but has minor misalignment issues. The $w^F$, $w^V$ and $w^T$ indicate the fusion, visible and thermal modality weight, respectively.}
  \label{fig:Sup_MPAW}
\end{figure*}

\begin{table}[t]
     \renewcommand{\tabcolsep}{2mm}
   \centering
   \begin{center}
      \resizebox{0.999\linewidth}{!}
      {
         \begin{tabular}{c c ccc}
            \Xhline{3\arrayrulewidth}
             \bf \multirow{2}{*}{Methods} &\bf Threshold & \multicolumn{3}{c}{\bf 30\%}\\ \cmidrule(lr){2-2}\cmidrule(lr){3-5}
             & \bf Pos. / Neg. & \bf All & \bf Day & \bf Night\\\hline
            Supervised & -  & 13.89 & 15.50 & 10.88 \\ \cdashline{1-5}
            \multirow{6}{*}{\bf SAMPD (Ours)}&0.9 / 0.9 & 10.62 & 11.71 & 8.59  \\
            &0.9 / 0.8 & 10.47 & 12.12 & 6.61  \\
            &0.9 / 0.7 & \bf 8.56 &\bf 10.55 &\bf 5.62 \\
            &0.9 / 0.6 &  10.00 & 11.60 &   6.75 \\
            &0.8 / 0.7 &  \underline{9.95} &  \underline{11.54} &  6.68 \\
            &0.8 / 0.6 &  10.05 &  11.89 & \underline{6.44} \\
            \Xhline{3\arrayrulewidth}
            
         \end{tabular}
        
      }
   \end{center}
   
   \caption{Analysis of thresholds (\textit{i.e.,} $\tau_1$, $\tau_2$) of the PPE module on KAIST dataset. We adopt Miss Rate (MR) as an evaluation metric. We select the thresholds ($\tau_1$ / $\tau_2$) as `0.9/0.7' in our main paper, which shows the highest performances. \textbf{Bold}/\underline{underlined} fonts indicate the best/second-best results. }
   \vspace{-0.2cm}
\label{table:threshold}
\end{table}
\begin{table}[t]
    \renewcommand{\tabcolsep}{3.0mm}
    \centering
    \resizebox{0.999\linewidth}{!}{
    \begin{tabular}{c c c c c c c c}
        \Xhline{3\arrayrulewidth}
        \rule{0pt}{10pt}
        \multirow{2}{*}{\bf Methods} & \multicolumn{3}{c}{\bf KAIST} \\ 
        \cmidrule(lr){2-4}
        &\bf  All         &\bf  Day         & \bf Night \\\hline
        \rule{0pt}{9.0pt}
        Limited data (10\%)     
        & 12.68     & 14.23     & 9.54  \\
        Sparse annotation (30\%)  
        & 13.89     & 15.50     & 10.88 \\\Xhline{3\arrayrulewidth}
    \end{tabular}
    }
    \caption{Detection results (MR) of our SAMPD with 30\% removal percentage and 10\% limited data setting with randomly selected 10\% labels on KAIST dataset. The 10\% (Limited data) and 30\% (Sparse annotation) is the most commonly used setting in each scenario, respectively.}

\label{table:semi_sparse}
\end{table}


\section{Analysis of the PPE Module Threshold}
\label{sec:3}
We investigate the performance of our method by varying the PPE module thresholds ($\tau_1$ and $\tau_2$). As shown in Table \ref{table:threshold}, even we change the thresholds, our SAMPD outperforms the `Supervised' method and all previous SAOD methods (see Table 1 in the main paper). When $\tau_1=0.9$ and $\tau_2=0.7$, our SAMPD achieves the best performance. When considering uncertain samples with ($\tau_1$/$\tau_2$) values of (0.9/0.8), (0.9/0.7), (0.9/0.6), the performance is higher compared to the case where uncertain samples are not considered during training (0.9/0.9). It indicates that by considering the uncertain samples, we can enhance the overall performance. 


\section{Sparsely Annotated vs. Limited Data}
\label{sec:4}
To investigate how challenging sparsely annotated situations are, we further conduct a comparative analysis between our method under sparse annotation and the limited data scenario on the KAIST dataset. To this end, we set the 10\% randomly selected labeled image for the limited data scenario and 30\% for the sparsely annotated scenario, because this is the most commonly used setting in limited data (\textit{e.g.,} semi-supervised scenario \cite{wang2023consistentteacher}) and sparsely annotated object detection environments \cite{wang2023calibratedteachersparselyannotated}. The sparsely annotated scenario, including 70\% of the annotations with only 30\% removed, looks relatively easier compared to the scenario using only 10\% of the data. However, as shown in Table \ref{table:semi_sparse}, despite having a greater quantity of annotations, the sparsely annotated scenario shows inferior performance with 1.21 MR for `All' setting. It indicates that even with a higher quantity of annotations in the sparsely annotated environment, performances can be reduced if annotations for pedestrians are not fully represented across all image pairs. Moreover, since the multispectral domains contain visible and thermal images, a reduction in annotation density leads to a more significant drop in performance. It demonstrates that our sparsely-annotated scenario presents more challenges compared to limited data scenarios, emphasizing their relevance in real-world scenarios.

\section{SAMPD on advanced multispectral pedestrian detection (MPD) models}
The SSD is widely adopted in the MPD task (e.g., MLPD, ProbEn, MBNet) for visible, thermal, and combined modalities. Specifically, our approach utilizes the same SSD+VGG16 architecture as MLPD, serving as its foundation.

To evaluate the generalization ability of our framework with sparsely annotated labels (30\% sparsely), experiments were conducted on two advanced models, MLPD and ProbEn. As shown in Table \ref{table:advanced_SAMPD}, the proposed SAMPD consistently enhances performance on these models, demonstrating its robust generalization capability.

\section{Compare with other MPD models}
As shown in the table \ref{table:comp_30_100}, in a 30\% sparsely annotated setting (30\% sparsely), SAMPD shows comparable performances to existing recent MPD models (e.g., MLPD, ProbEn, MBNet) that use the full labels (100\% label). In the fully labeled setting, our SAMPD outperforms all the other models. This demonstrates that our method is effective in both sparse and fully labeled (but potential incompleteness may exist) settings.

\section{Computational Costs}
Since our SAMPD adopts a teacher-student architecture, training time slightly increases with minimal parameter overhead. Specifically, the training time for the baseline model is 0.483 seconds with 227.8M parameter, while our SAMPD takes 0.512 seconds (6\% increase) with 236.7M (4\% increase). However, inference time remains unchanged at 0.014 seconds per image with 227.8M parameters, as we only use the student model.

\begin{table}[t]
    \renewcommand{\tabcolsep}{3.0mm}
    \centering
    \resizebox{0.999\linewidth}{!}{
    \begin{tabular}{c c c c c c c c}
        \Xhline{3\arrayrulewidth}
        \rule{0pt}{10pt}
        \multirow{2}{*}{\bf Methods} & \multicolumn{3}{c}{\bf KAIST} \\ 
        \cmidrule(lr){2-4}
        &\bf  All         &\bf  Day         & \bf Night \\\hline
        \rule{0pt}{9.0pt}
        MLPD           & 13.89     & 15.50      & 10.88    \\
        MLPD + SAMPD   & \bf 8.56  & \bf 10.55	& \bf 5.62 \\
        ProbEn         & 11.54     & 13.00      & 8.15     \\
        ProbEn + SAMPD & \bf 7.75  & \bf 8.90   & \bf 5.65 \\\Xhline{3\arrayrulewidth}
    \end{tabular}
    }
    \caption{Detection results (MR) of our SAMPD on advanced multispectral pedestrian detection models with 30\% removal percentage on KAIST dataset. \textbf{Bold} font indicates the best results.}
\label{table:advanced_SAMPD}
\end{table}

\begin{table}[t]
    \renewcommand{\tabcolsep}{3.0mm}
    \centering
    \resizebox{0.999\linewidth}{!}{
    \begin{tabular}{c c c c c c c c c}
        \Xhline{3\arrayrulewidth}
        \rule{0pt}{10pt}
        \multirow{2}{*}{\bf Methods} & \multirow{2}{*}{\bf Setting} & \multicolumn{3}{c}{\bf KAIST} \\ 
        \cmidrule(lr){3-5}
        & &\bf  All         &\bf  Day         & \bf Night \\\hline
        \rule{0pt}{9.0pt}
        MLPD + SAMPD    & 30\% sparsely   & 8.56       & 10.55     & 5.62  \\
        ProbEn + SAMPD  & 30\% sparsely   & 7.75	   & 8.90      & 5.65  \\
        MBNet           & 100\% label     & 8.13	   & 8.28	   & 7.86  \\
        MLPD            & 100\% label     & 7.58	   & 6.95	   & 7.95  \\
        ProbEn          & 100\% label     & 6.76	   & 7.81	   & \bf 5.02  \\
        MLPD + SAMPD    & 100\% label     & 6.50	   & 6.85	   & 5.99  \\
        ProbEn + SAMPD	& 100\% label     & \bf 5.94   & \bf 6.41  & 5.06 \\\Xhline{3\arrayrulewidth}
    \end{tabular}
    }
    \caption{Detection results (MR) of our SAMPD with 30\% removal percentage and fully annotated setting on KAIST dataset. We compared with existing recent MPD models. \textbf{Bold} font indicates the best results.}

\label{table:comp_30_100}
\end{table}




\begin{figure*}[t] 
  \centering
  \includegraphics[width=\linewidth]{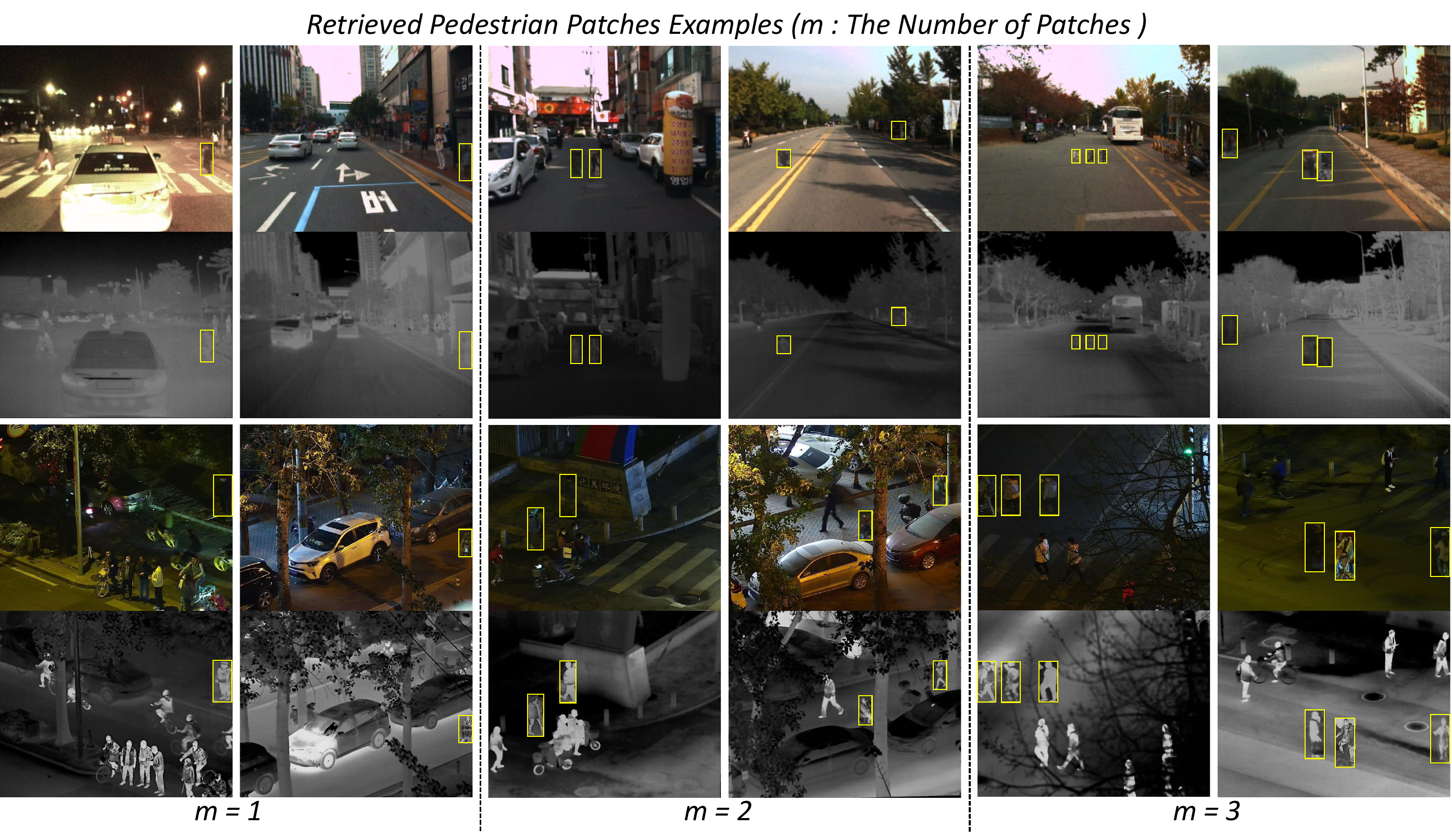}
  \caption{Examples of our APRA module. Yellow bounding-boxes denote the example of the pedestrian patch. Note that, $m$ is the number of pedestrian patches and the range of $m=1$ to $m=3$.}
  \label{fig:Sup_More_APRA}
\end{figure*}

\begin{figure*}[t] 
  \centering
  \includegraphics[width=\linewidth]{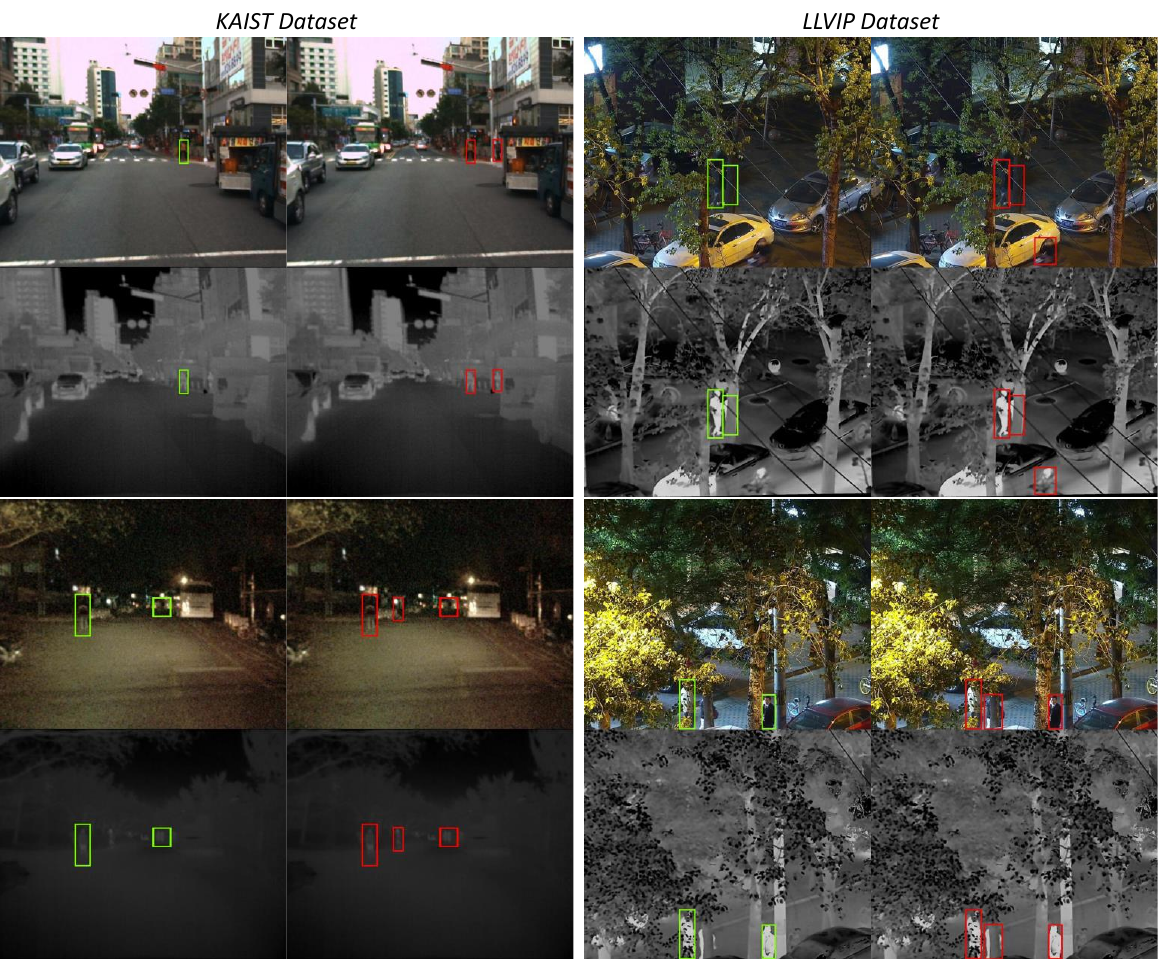}
      \caption{Visualization results of refined ground-truth in fully annotated scenario on KAIST and LLVIP datasets. Each image pair indicates visible and thermal images, respectively. Green bounding-boxes denote the original ground-truth (`Static' approach) and red bounding-boxes indicate the refined ground-truth (`Dynamic' approach).}
  \label{fig:Fully_Dynamic}
\end{figure*}
\begin{figure*}[t] 
  \centering
  \includegraphics[width=\linewidth]{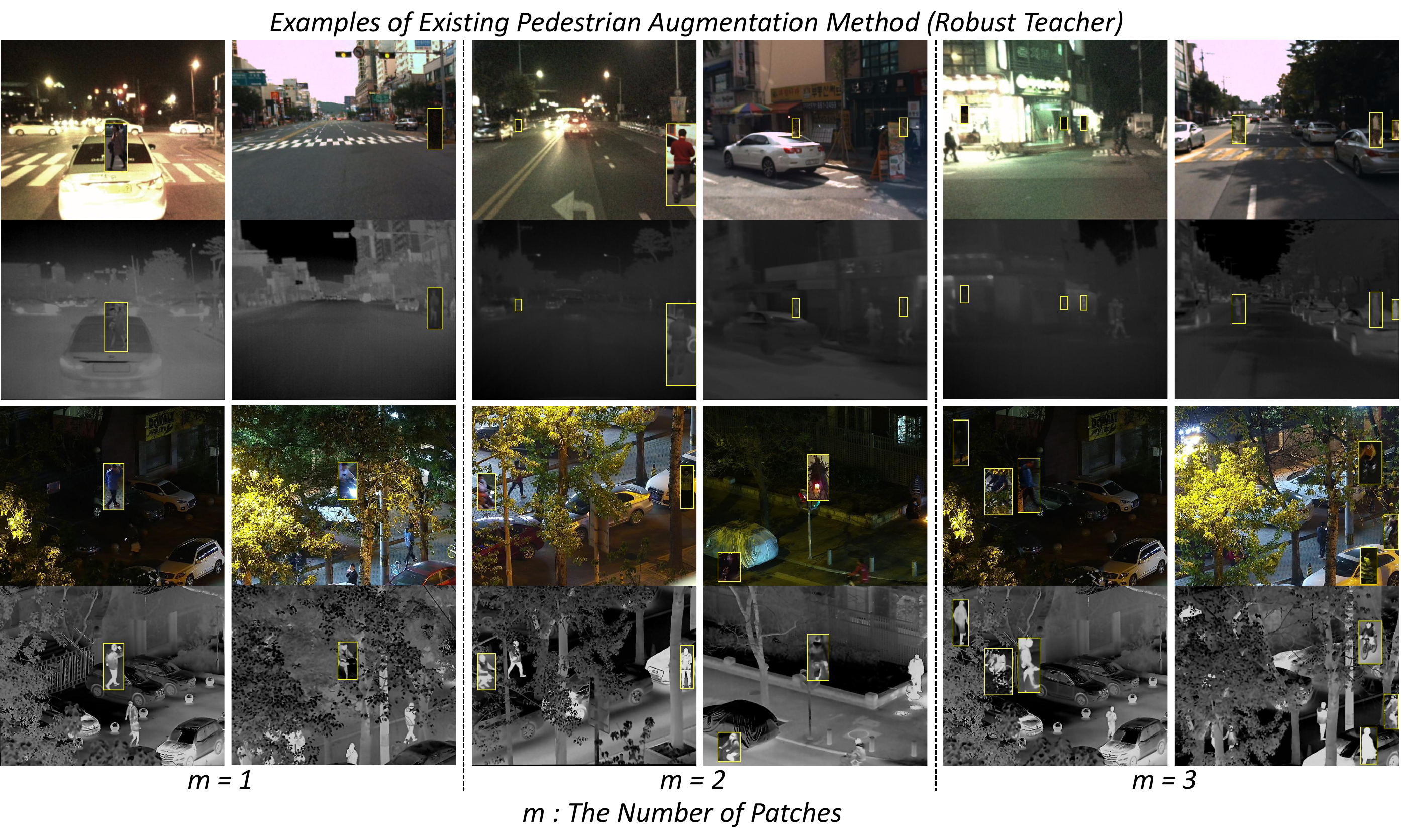}
  \caption{Examples of existing pedestrian augmentation method (Robust Teacher). Yellow bounding-boxes denote the example of the pedestrian patch. Note that, $m$ is the number of pedestrian patches and the range of $m=1$ to $m=3$.}
  \label{fig:Sup_More_CVIU}
\end{figure*}

\section{More Visualization Results}
\label{sec:5}
\subsection{Effect of the MPAW Module}
We additionally provide examples of the pedestrian detection results of our multispectral pedestrian-aware adaptive weight (MPAW) module. MPAW module compares the ground truth pedestrian feature map and the pseudo-label pedestrian feature map to verify the quality of the pseudo-label. Note that, when the quality of the pseudo-label is high, the value of cosine-similarity will be high and highly reflects to model training. Conversely, when the quality of the pseudo-label is low, the value of cosine-similarity will be small. We show the visualization results when the pseudo-label quality is determined to be high and low. The results are shown in Figure \ref{fig:Sup_MPAW}. The `middle quality' indicates that pseudo-labels capture the pedestrians, but the bounding-box has some uncertainty (\textit{e.g.,} slightly misaligned bounding-box).
By following this manner, we determine the threshold of our PPE module. We set the threshold for positive pseudo-labels as greater than or equal to 0.9 for each pseudo-label due to almost of these pseudo-labels case capture the pedestrians correctly. Conversely, we set the threshold for negative pseudo-labels as less than or equal to 0.7 for each pseudo-label due to almost of these pseudo-label cases failed to capture the pedestrian correctly. Furthermore, the middle-quality pseudo-labels, which correctly identify the pedestrian but have minor alignment issues, were also considered.

\subsection{Visualization Results of the our APRA Module}
\label{sec:6}
First, we present additional visualization results of the adaptive pedestrian retrieval augmentation (APRA) module by varying the number of the $m$ pedestrian patches. The results are shown in Figure \ref{fig:Sup_More_APRA}. To enhance comprehension, we limit the maximum number of patches $m$ generated by our APRA module to 3. The results indicate that pedestrian patches generated by our APRA module appear natural and blend smoothly with both visible and thermal images. Second, we present the `Dynamic' approach on fully annotated scenario. This means that the data that has been artificially removed does not exist at all. In Figure \ref{fig:Fully_Dynamic}, there is still sparse annotation, and our method effectively detects and converts the missing annotation to ground-truth.

\subsection{Visualization Comparisons of Existing Pedestrian Augmentation Method}
\label{sec:6}
In Figure \ref{fig:Sup_More_CVIU}, we present the visualization results of the existing pedestrian augmentation method (Robust Teacher) \cite{LI2023103788} to highlight the differences between the our APRA module and existing method. Unlike APRA module, the existing method attaches pedestrian patches exactly where they are in the cropped image. However, this often leads to unnatural insertions, such as pedestrians appearing on top of or in front of cars, or in other implausible locations based on human perception. Moreover, the method does not adaptively adjust the patches, resulting in mismatches like daytime pedestrian patches being inserted into nighttime scenes or vice versa. These unnatural insertions can lead to noise, as the boundaries or locations of the patches are not consistent with the surrounding context. Therefore, we believe that this may be the reason why the method does not perform as well as APRA module.

\subsection{Comparison with SAOD Methods and Ours}
\label{sec:7}
\begin{figure*}[t] 
  
  \centering
  \includegraphics[width=\linewidth,height=20cm, keepaspectratio]{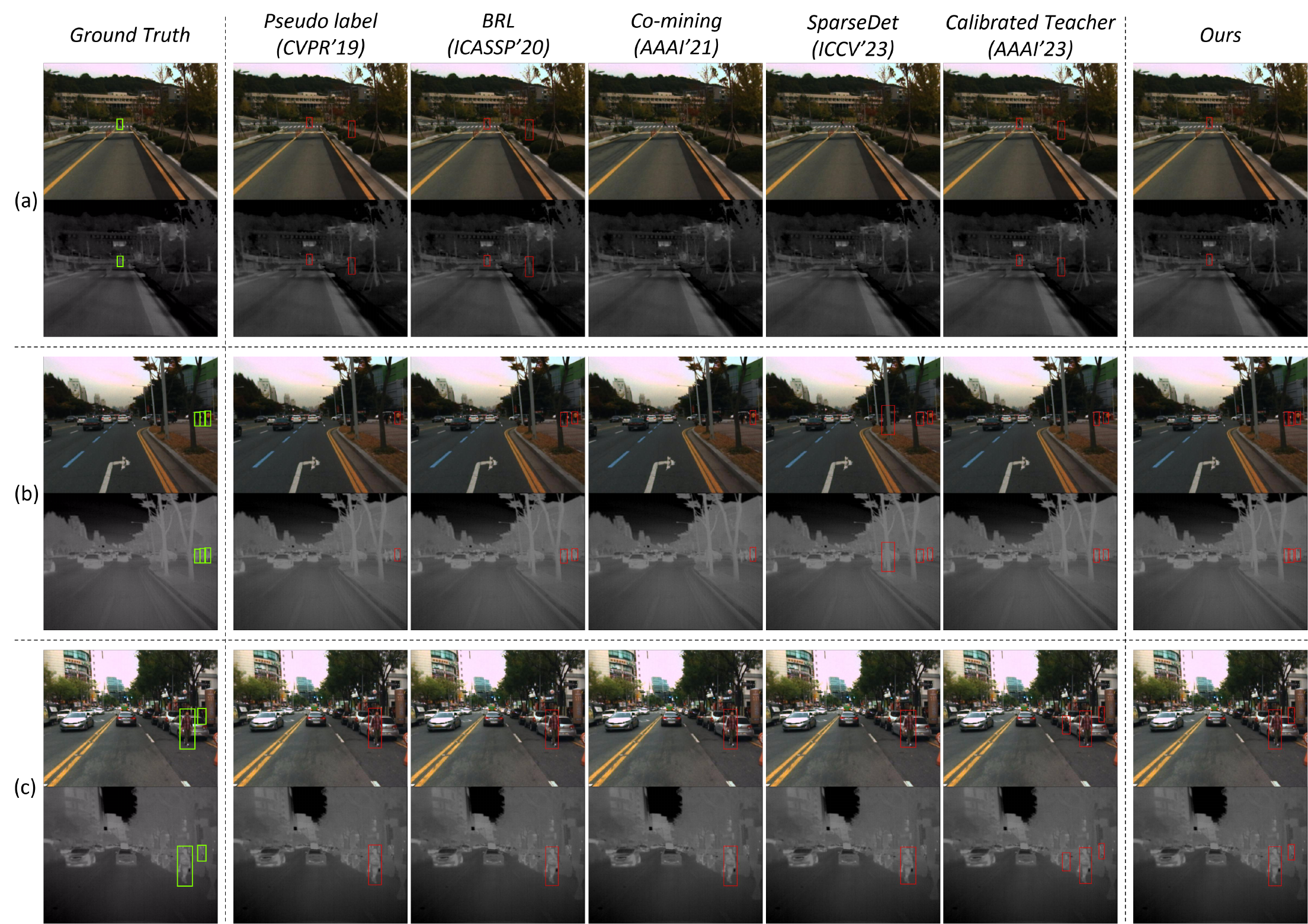}
  \caption{Detection results of the proposed method with existing SAOD methods on KAIST dataset (Green: ground-truth bounding-box and red: predicted bounding-box). Each image pair indicates visible and thermal images, respectively. (a), (b), and (c) represent the removal percentages of 30\%, 50\%, and 70\%, respectively.}
  \label{fig:Sup_kaist}
\end{figure*}

\noindent \textbf{KAIST Dataset.} We provide visualization results comparing the proposed method with the existing SAOD methods \cite{niitani2019sampling,zhang2020solving,wang2021comining,wang2023calibratedteachersparselyannotated,suri2023sparsedet} on the KAIST dataset \cite{hwang2015multispectral}, by varying the removal percentages (30\%, 50\%, 70\%). The results are shown in Figure \ref{fig:Sup_kaist}. When the pedestrians are occluded or cropped, the previous SAOD methods are struggle to the detection, resulting in performance degradation (see Table P1 and Table P2 of our main paper). In contrast, our method can detect pedestrians in such challenging cases. This improvement is attributed to (1) the effectiveness of the MPAW module in reducing the impact of low-quality pseudo-labels, (2) the ability of the PPE module to direct the teacher model in generating high-quality pseudo-labels, and (2) the contribution of the APRA module in utilizing information from a diverse range of samples. As a result, these elements enable our SAMPD to learn robustly even with sparsely annotated labels in the multispectral domain. \\

\begin{figure*}[t] 
  
  \centering
  \includegraphics[width=\linewidth,height=20cm, keepaspectratio]{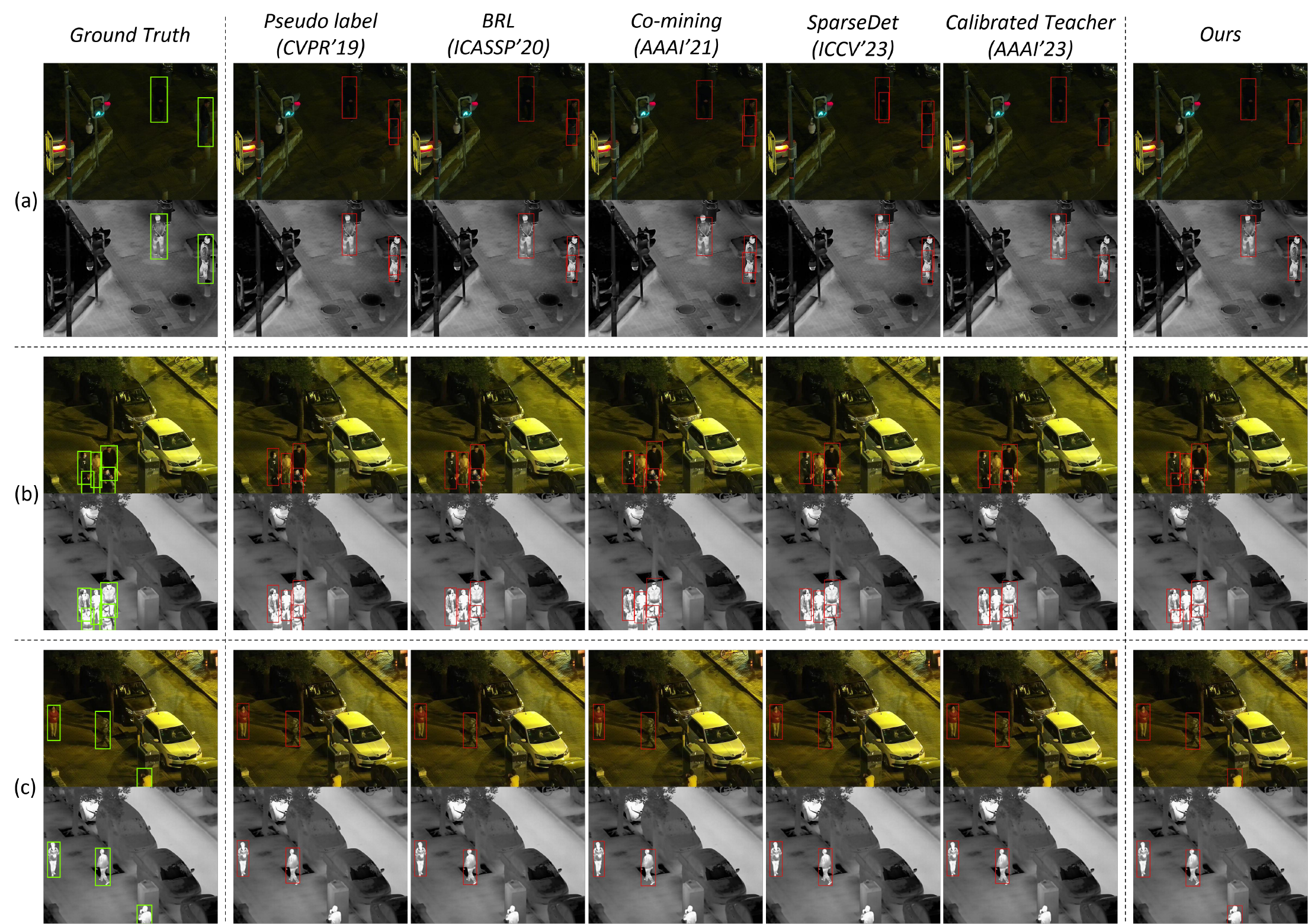}
  \caption{Detection results of the proposed method with existing SAOD methods on LLVIP dataset (Green: ground-truth bounding-box and red: predicted bounding-box). Each image pair indicates visible and thermal images, respectively. (a), (b), and (c) represent the removal percentages of 30\%, 50\%, and 70\%, respectively.}
  \label{fig:Sup_llvip}
\end{figure*}

\noindent \textbf{LLVIP Dataset.} In addition, Figure \ref{fig:Sup_llvip} shows the visualization of the detection results of our method and the existing SAOD methods \cite{niitani2019sampling,zhang2020solving,wang2021comining,wang2023calibratedteachersparselyannotated,suri2023sparsedet} on the LLVIP dataset\cite{jia2021llvip}. As shown in the figure, similar to the results observed with the KAIST dataset (Figure \ref{fig:Sup_kaist}), our method successfully detects occluded and cropped pedestrian samples in the LLVIP dataset. Unlike our methods, the existing methods \cite{niitani2019sampling,zhang2020solving,wang2021comining,wang2023calibratedteachersparselyannotated,suri2023sparsedet} show a significant number of misaligned predictions. This demonstrates that our method effectively utilizes multispectral knowledge to achieve robust learning for the sparsely-annotated scenarios.

\subsection{Effect of the Proposed Modules}
\label{sec:9}

\begin{figure*}[t] 
  \centering
  \includegraphics[width=0.8\linewidth]{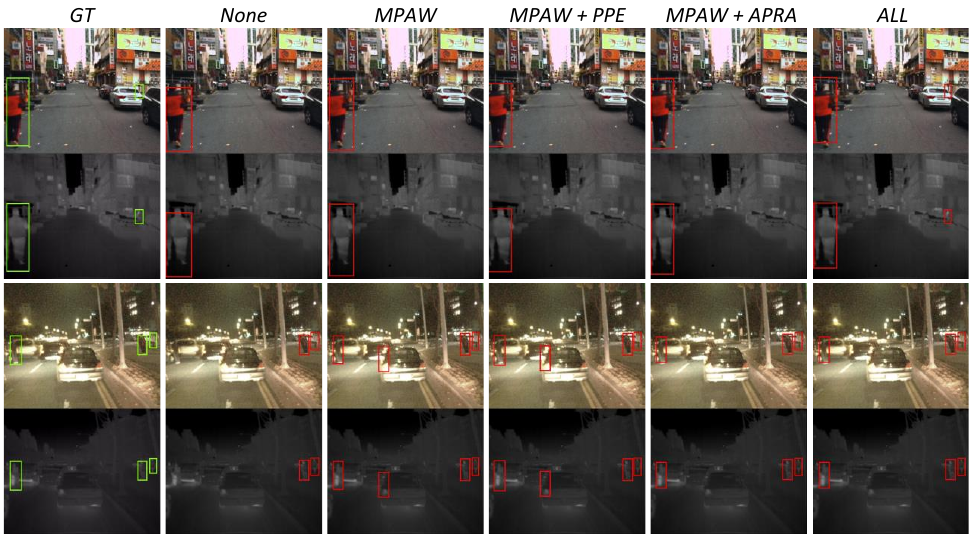}
  \caption{Detection results of the proposed method on KAIST dataset (Green: ground-truth bounding-box and red: predicted bounding-box). Each image pair indicates visible and thermal images, respectively. `GT’ means ground truth, `MPAW’ denotes multispectral pedestrian-aware adaptive weight module, `PPE’ indicates positive pseudo-label enhancement module, and `APRA’ denotes adaptive pedestrian retrieval augmentation module.}
  \label{fig:Sup_Abl}
\end{figure*}

We provide additional visualization examples of our proposed modules (\textit{i.e.,} MPAW module, PPE module, APRA module) on the KAIST dataset. The results are shown in Figure \ref{fig:Sup_Abl}. Experiments are conducted with 30\% of sparsely annotated scenarios. `None' indicates the results of the baseline detector without our method. Adding MPAW module (`MPAW') leads to improving the detection results. Additionally, utilizing modules such as PPE module and APRA module significantly improves the precision of the detection outcomes. When all three modules are considered, we achieve the best visualization results.


\section{References}
\noindent Hwang, S.; Park, J.; Kim, N.; Choi, Y.; and So Kweon, I. 2015. Multispectral pedestrian detection: Benchmark dataset and baseline. In \textit{Proceedings of the IEEE conference on computer vision and pattern recognition}, 1037–1045.

\noindent Jia, X.; Zhu, C.; Li, M.; Tang, W.; and Zhou, W. 2021. LLVIP: A visible-infrared paired dataset for low-light vision. In \textit{Proceedings of the IEEE/CVF international conference on computer vision}, 3496–3504.

\noindent Li, S.; Liu, J.; Shen, W.; Sun, J.; and Tan, C. 2023. Robust Teacher: Self-correcting pseudo-label-guided semi-supervised learning for object detection. \textit{Computer Vision and Image Understanding}, 235: 103788.

\noindent Niitani, Y.; Akiba, T.; Kerola, T.; Ogawa, T.; Sano, S.; and Suzuki, S. 2019. Sampling Techniques for Large-Scale Object Detection from Sparsely Annotated Objects. arXiv:1811.10862.

\noindent Suri, S.; Rambhatla, S. S.; Chellappa, R.; and Shrivastava, A. 2023. SparseDet: Improving Sparsely Annotated Object Detection with Pseudo-positive Mining. arXiv:2201.04620. 

\noindent Wang, H.; Liu, L.; Zhang, B.; Zhang, J.; Zhang, W.; Gan, Z.; Wang, Y.; Wang, C.; and Wang, H. 2023a. Calibrated Teacher for Sparsely Annotated Object Detection. arXiv:2303.07582.

\noindent Wang, T.; Yang, T.; Cao, J.; and Zhang, X. 2021. Co-mining: Self-Supervised Learning for Sparsely Annotated Object Detection. arXiv:2012.01950. 

\noindent Wang, X.; Yang, X.; Zhang, S.; Li, Y.; Feng, L.; Fang, S.; Lyu, C.; Chen, K.; and Zhang, W. 2023b. ConsistentTeacher: Towards Reducing Inconsistent Pseudo-targets in Semi-supervised Object Detection. arXiv:2209.01589.

\noindent Zhang, H.; Chen, F.; Shen, Z.; Hao, Q.; Zhu, C.; and Savvides, M. 2020. Solving Missing-Annotation Object Detection with Background Recalibration Loss. arXiv:2002.05274.

\end{document}